\documentclass[10pt,twocolumn,letterpaper]{article}

\pdfoutput=1

\usepackage{cvpr}              

\usepackage{tcolorbox,titletoc}
\usepackage{xstring,catchfile}
\usepackage[accsupp]{axessibility}
\definecolor{cvprblue}{rgb}{0.21,0.49,0.74}
\usepackage[pagebackref,breaklinks,colorlinks,allcolors=cvprblue,hypertexnames=false]{hyperref}
\usepackage[table,x11names]{xcolor} 
\usepackage[accsupp]{axessibility}

\usepackage{amsmath,amsfonts,bm}

\def\eqref#1{equation~\ref{#1}}

\def\1{\bm{1}}

\DeclareMathAlphabet{\mathsfit}{\encodingdefault}{\sfdefault}{m}{sl}
\SetMathAlphabet{\mathsfit}{bold}{\encodingdefault}{\sfdefault}{bx}{n}

\usepackage{makecell, graphicx, romannum, amssymb, amsmath, ccaption, setspace, algorithm, algpseudocode, enumitem, pifont, etoolbox, fontawesome5, longtable, array, arydshln, bbding}

\def\paperID{****} 
\def\confName{CVPR}
\def\confYear{2026}

\title{ReFoCUS: Reinforcement-guided Frame Optimization\\for Contextual Understanding}

\author{
Hosu Lee$^{1}$\thanks{Equal contribution. $\dagger$ Corresponding author.} \quad\quad
Junho Kim$^{2}$\footnotemark[1] \quad\quad
Hyunjun Kim$^{1}$ \quad\quad
Yong Man Ro$^{1\dagger}$ \\
\small$^{1}$Korea Advanced Institute of Science \& Technology \quad $^{2}$University of Illinois Urbana-Champaign \\
{\tt\small \{leehosu01,kimhj709,ymro\}@kaist.ac.kr, arkimjh@illinois.edu}
}

\begin{document}
\pagenumbering{arabic}

\maketitle

\begin{abstract}
Recent progress in Large Multi-modal Models (LMMs) has enabled effective vision-language reasoning, yet the ability to video understanding remains constrained by suboptimal frame selection strategies, albeit with the rapid development of video-specialized LMMs. Prior works attempted to solve this with static heuristics or external retrieval modules to feed frame-level information, but these approaches often fail to capture visual cues grounded to the given user queries conflating raw visual dynamics with true semantic relevance. In this paper, we introduce \textbf{ReFoCUS} (\textbf{Re}inforcement-guided \textbf{F}rame \textbf{O}ptimization for \textbf{C}ontextual \textbf{U}nder\textbf{S}tanding), the first framework to integrate online policy-gradient reinforcement learning into frame-level optimization for video-LLMs. \textbf{ReFoCUS} aims to learn a frame selection policy, leveraging reward signals derived from reference models to capture their underlying scoring behavior over frame combinations that best support temporally grounded responses. To efficiently explore the large combinatorial frame space, we employ an autoregressive and query-conditional selection architecture that ensures contextual consistency while reducing complexity. Our policy learning removes the need for explicit frame-level supervision, as it implicitly discovers optimal and semantically consistent frame compositions. \textbf{ReFoCUS} consistently improves reasoning accuracy across multiple video QA benchmarks, demonstrating the advantage of aligning frame selection with model-internal utility.
\end{abstract}

\section{Introduction}
After the wide adoption of Large Language Models (LLMs)~\cite{brown2020language,touvron2023llama,yang2024qwen2} across language-centric applications, users can now engage with multi-modal systems~\cite{chatgpt,gpt4,reid2024gemini} through back-and-forth conversations, marking the beginning of Large Multi-modal Model (LMM) era~\cite{gpt4v, gpt4o}. As the exceptional perception and reasoning capabilities of LLMs have rapidly advanced through large-scale web-scraped corpora, the emergence of high-quality multi-modal paired datasets~\cite{bain2021frozen, xue2022advancing, chen2024panda} has enabled LMMs to achieve cross-modal consistency through alignment pre-training, followed by post-training stages such as supervised fine-tuning and reinforcement learning-based preference optimization. Accordingly, various LMMs~\cite{liu2023improved,li2024llavanext, huang2024language} can seamlessly process linguistic and visual information with robust performance across diverse vision-language tasks.

\begin{figure}[t]
\centering
\includegraphics[width=1.0\linewidth]{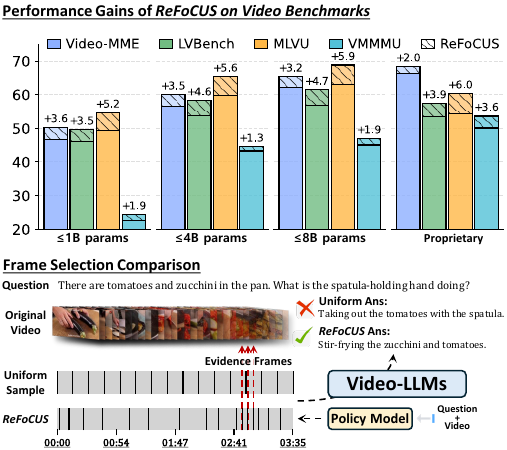}
\vspace{-8mm}
\caption{
Overall performance gains of \textit{ReFoCUS} across benchmarks and model scales, exemplifying semantically relevant frame selection over uniform sampling.
}
\vspace{-5mm}
\label{fig:1}
\end{figure}

Even if pioneer works~\cite{alayrac2022flamingo, chen2023internvl, li2023blip, liu2023visual} have successfully developed numerous LMMs, achieving impressive performance through visual modality integration, the understanding of video content remains substantially below human-level capability. A major limitation arises because many existing video-LLMs~\cite{maaz2023video, lin2023video, lin2024vila} employ on simplistic strategies to input spatio-temporal dynamics, treating a video as a sequence of static images (\textit{e.g.,} uniform frame sampling). Furthermore, constrained by the limited context length of backbone language models, these systems struggle to maintain inter-modal alignment and continuity over extended temporal sequences, leading to suboptimal contextual understanding especially in complex or long-form videos.

To address this limitation, several works~\cite{lin2023video,li2024mvbench,xu2024pllava} have delved into adaptive frame sampling strategies to align frame-level information and deliver relevant visual cues into the models. Beyond uniform frame sampling, many studies selectively retrieve relevant video segments using auxiliary retrieval modules~\cite{kim2024salova,jeong2025videorag} or memory-augmented strategies~\cite{song2024moviechat,he2024ma}, but these approaches remain dependent on external modules and are not jointly optimized with the model’s internal reasoning process. More recent works~\cite{liu2025bolt,sun2025mdp3,zou2025air} have introduced training-free search algorithms to identify informative frames, which are incorporated into pre-trained video-LLMs as prior cues. While these methods alleviate redundant frame selection, they still struggle to identify frame compositions contextually coherent and temporally grounded to the user queries, often conflating raw visual dynamics with true semantic relevance.

We introduce \textbf{\textit{ReFoCUS}} (\textbf{Re}inforcement-guided \textbf{F}rame \textbf{O}ptimization for \textbf{C}ontextual \textbf{U}nder\textbf{S}tanding), the first framework to integrate online policy-gradient reinforcement learning (RL) into frame-level optimization for video-LLMs. While previous works have mainly applied RL into multi-modal systems by optimizing output behaviors~\cite{wang2024mdpo,ahn2024tuning,zhou2024aligning} (\textit{e.g.,} aligning generated responses with human preferences), its potential for visual-level optimization remains largely unexplored. In this work, we leverage online RL to train a policy model that interactively selects frames most relevant to the user’s query, guided by reward signals derived from reference models that capture their internal scoring behavior over visual evidence. As summarized in~\cref{fig:1}, unlike heuristic selection, \textit{ReFoCUS} adaptively focuses on semantically and temporally coherent visual contexts, enhancing reasoning performance across complex, event-rich video scenarios with substantial performance gains.

To achieve our RL-based policy learning, we handle the following two main challenges: 
(\lowercase\expandafter{\romannumeral1}) collecting frame-level supervision is significantly resource-intensive and infeasible compared to textual preference data, due to the combinatorial explosion inherent in lengthy videos. To address this, we employ reference models to evaluate sampled frame subsets, enabling group-wise relative reward modeling across candidates and guiding the policy optimization through an effective advantage function and
(\lowercase\expandafter{\romannumeral2}) during online policy-gradient optimization, the extensive frame-level search space makes it difficult for the policy to efficiently explore and identify optimal frame subsets. Accordingly, we introduce an autoregressive, conditional frame selection architecture for \textit{ReFoCUS}, which progressively identifies query-relevant frames conditioned on previously selected contexts, which effectively reduces the frame search complexity while ensuring contextual and temporal consistency.

Our framework is lightweight and model-agnostic, seamlessly integrating with diverse video-LLMs and yielding consistent and significant gains across video QA benchmarks, thereby corroborating its efficacy in enhancing contextually grounded visual reasoning.

Our contributions can be summarized as follows:
\begin{itemize}
    \item We propose \textit{ReFoCUS}, online policy-gradient reinforcement learning framework for direct frame-level optimization, enabling video-LLMs to internalize their own visual scoring behavior and enhance contextually grounded video reasoning through frame-level optimization.
    \item We design an autoregressive, conditional frame selection architecture that efficiently explores the large combinatorial frame space by progressively selecting query-relevant frames based on prior context, ensuring temporal and semantic consistency throughout the selection process.
    \item We demonstrate that \textit{ReFoCUS} consistently improves reasoning accuracy across multiple video QA benchmarks, validating its generality and practical effectiveness in optimizing visual evidence for contextual understanding.
\end{itemize}

\section{Related Work}
\paragraph{Large Multi-modal Models for Video Understanding.}
As LLMs~\cite{brown2020language, touvron2023llama} have advanced, multi-modal integration has led to the emergence of LMMs~\cite{zhu2023minigpt, chen2023internvl, dong2024internlm} capable of processing both visual and textual inputs. Building on foundational multi-modal instruction tuning~\cite{liu2023visual, ye2023mplug, dai2023instructblip}, recent works have specifically expanded their scope towards video modality~\cite{lin2023video, maaz2023video, xu2024pllava}, aiming to achieve deeper spatio-temporal reasoning and better contextual understanding. Several key directions include developing enhanced temporal modeling strategies~\cite{xu2024pllava, song2024moviechat}, refining alignment mechanisms across modalities~\cite{cha2023honeybee, mckinzie2024mm1}, and leveraging larger, higher-quality video-text paired datasets~\cite{li2024llavanext, dong2024internlm}. Despite such advances, many video-LMMs still rely on sparse frame sampling or limited temporal windows, constraining their ability to dynamically capture visual-level semantics within a video. Recent approaches attempt to address these limitations through techniques such as memory augmentation~\cite{he2024ma}, extended context modeling~\cite{zhang2024long}, or frame-selection. However, both training-based~\cite{yu2025framevoyager,tang2025tspo,yao2025k,ye2025re,hu2025m,yu2023self} and heuristic~\cite{wang2024videoagent,liu2025bolt,sun2025mdp3,zou2025air} frame-selection methods typically align frames or regions to the query but sample them independently, without conditioning on previously chosen evidence. In contrast, our approach performs autoregressive selection, sequentially gathering semantically coherent visual evidence for better temporal reasoning.

\vspace{-4mm}
\paragraph{RL-based Post-training Preference Optimization.} Recent research has increasingly integrated RL into LMMs as a method for post-training preference optimization. Initial efforts~\cite{ouyang2022training,sun2023aligning,yu2024rlhf} primarily targeted mitigating hallucinations and enhancing factual accuracy in model responses through RL with human feedback (RLHF), with a particular emphasis on learning from pairwise preference comparisons over textual outputs. Advancing beyond PPO-style RLHF pipelines, Direct Preference Optimization (DPO)~\cite{rafailov2023direct} formulates preference alignment more directly through a supervised objective, and has been further adapted to multi-modal settings for aligning outputs with human preferences~\cite{wang2024mdpo, zhou2024aligning}. However, existing approaches predominantly focus on aligning textual outputs with human preferences by updating the policy model accordingly, while paying little attention to the model’s visual input space. In contrast, our method, \textit{ReFoCUS}, shifts the focus to the input level by optimizing which visual content the model attends to. By leveraging reinforcement learning to identify informative frames, it aligns the model’s visual inputs with its internal scoring behaviors, enabling more coherent and context-aware video understanding.

\section{Proposed Method}
\begin{figure}[t]
\centering
\includegraphics[width=0.99\linewidth]{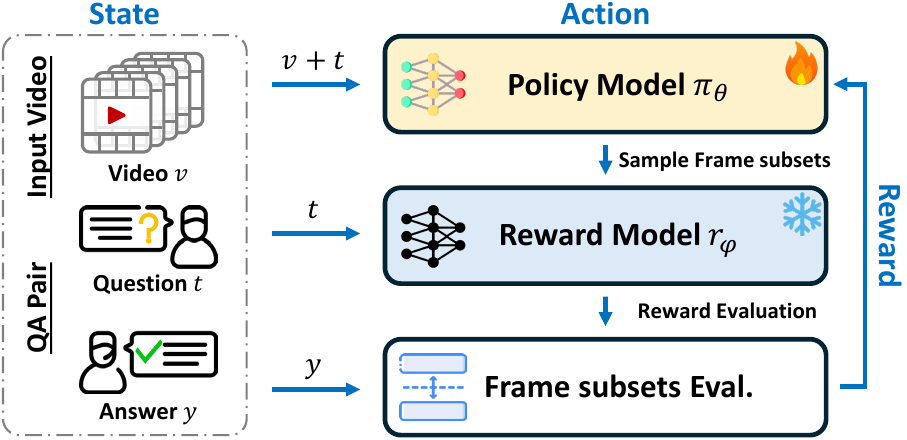}
\vspace{-5mm}
\caption{Overview of \textit{ReFoCUS} framework, illustrating policy-based frame selection guided by reward signals.}
\vspace{-4mm}
\label{fig:2}
\end{figure}

\paragraph{Overview.} We illustrate the overall \textit{ReFoCUS} pipeline in~\cref{fig:2}, highlighting its policy optimization process. Unlike existing approaches~\cite{rafailov2023direct, zhou2024aligning} that fine-tune LMMs to generate human-preferred textual outputs, \textit{ReFoCUS} extends online policy-gradient learning to the input-level, specifically, to frame selection in video-LLMs. The key idea is to align the model’s visual inputs with its internal scoring behavior, enabling the model to attend to the most informative spatio-temporal cues conditioned on the given query. That is, \textit{ReFoCUS} aims to learn a policy that samples an ideal frame combination that effectively guides the model toward accurate and consistent reasoning.

\textit{ReFoCUS} consists of two core components: learnable \textit{Policy Model} and frozen \textit{Reward Model}, which jointly facilitate learning toward contextually grounded frame selection. The policy model receives dense video sequences along with user queries and samples frame subsets that best support semantic relevance grounded to the queries. The reward model, in turn, serves as a reference evaluator for candidate subsets, translating its predictive confidence into reward feedback that drives efficient policy learning.

In detail, for a given video input $v$ and its corresponding QA pair $\langle t,y\rangle$, the policy model $\pi_{\theta}$ samples $N$ frame subsets. Each subset is then scored by the reference model $r_{\varphi}$, which estimates a reward based on its prediction confidence for $y$. These rewards update $\pi_{\theta}$ via policy-gradient optimization, encouraging it to progressively identify frame combinations that maximize the model’s own reward feedback and enhance query-grounded reasoning.

\subsection{Data Perspective for Frame-level Optimization}
\label{sec:data}
\paragraph{Learning without Frame-level Supervision.} We start from a data perspective. Unlike textual preference data, collecting frame-level supervision for video content is prohibitively expensive due to the combinatorial explosion of possible frame subsets. For instance, even with a moderate-length video containing $512$ candidate frames, selecting $32$ frames already yields ${\approx}10^{50}$ possible combinations, making annotation practically impossible. Moreover, it is infeasible to manually determine \textit{which frame subsets are truly informative or causally relevant to the model’s answers across arbitrary user queries}. Therefore, we instead leverage the output logits of $r_{\varphi}$ as feedback signals to incentivize $\pi_{\theta}$ to sample frame combinations that better align with high-confidence predictions.

To enable reliable reward estimation and maintain consistent evaluation quality from $r_{\varphi}$, we formulate our training environment as a multiple-choice QA setting with discrete answer options.
This closed-set setup provides a well-defined reward space, allowing the model to receive clear feedback signals, unlike open-ended QA, which often yields ambiguous reward distributions due to its inherently subjective nature. We collect a diverse and comprehensive pool of QA pairs from various video understanding datasets, including LLaVA-Video-178K~\cite{zhang2024video}, NExT-QA~\cite{xiao2021next}, ActivityNetQA~\cite{caba2015activitynet}, PerceptionTest~\cite{patraucean2024perception}, CinePile~\cite{rawal2024cinepile}, and VISTA-400K~\cite{ren2024vista}, resulting in a total of $962$K pairs.

\begin{figure*}[t]
\centering
\includegraphics[width=1.0\textwidth]{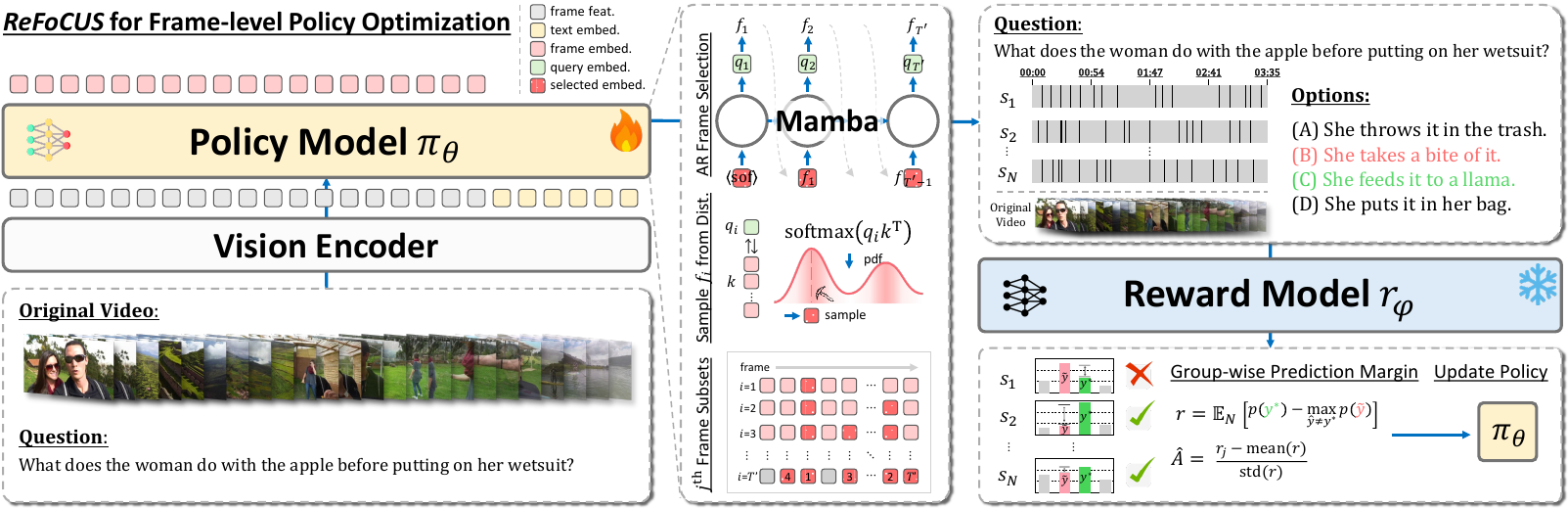}
\vspace{-8mm}
\caption{
Overview of \textit{ReFoCUS} framework. Given a video and query, $\pi_\theta$ autoregressively selects $N$ frame subsets, which are then scored by $r_\varphi$ based on their answer prediction margins. The resulting rewards guide our frame-level policy optimization.}
\vspace{-4mm}
\label{fig:3}
\end{figure*}

\vspace{-3mm}
\paragraph{Reward-variance Filtering Procedure.}
Online reinforcement learning critically depends on efficient data sampling strategies, as the quality and diversity of collected samples directly affect policy convergence~\cite{yu2025dapo}. In the case of video-LLMs, however, not all video-QA pairs contribute equally to learning. Certain instances produce nearly identical predictions across different frame subsets, resulting in flat reward landscapes and near-zero policy gradients when applying group-relative optimization~\cite{shao2024deepseekmath}. Such low-variance samples correspond to holistic or globally answerable questions that provide limited temporal supervision, thereby diluting the reinforcement signal, analogous to the gradient saturation phenomenon observed in preference optimization methods~\cite{ziegler2019fine, ouyang2022training}.

To address this inefficiency, we design a reward-variance filtering pipeline that prioritizes samples exhibiting strong temporal sensitivity. For each video $v$ with $T$ frames, we divide it into $8$ overlapping temporal windows with a fixed window size $w {=} \lceil T / 8 \rceil$ and stride $s {=} \lceil w / 2 \rceil$. Each window $W_i$ is paired with its complementary region $C_i {=} [0, T) \setminus W_i$, which covers the remaining portions of the video not included in $W_i$. From both $W_i$ and $C_i$, we uniformly sample $k {=} 32$ frames, yielding $16$ candidate subsets $\{c_{j}\}^{16}_{j=1}$. These subsets jointly capture both localized and contextual temporal evidence, enabling a comprehensive assessment of the model’s temporal sensitivity. (details in~\cref{sec:supp_data})

For each video–question pair $\langle v, t \rangle$, we compute the prediction results using a pretrained LMM~\cite{wang2024enhancing} on each frame subset, defined as the logit of the correct answer for that subset $c_j$. We then measure the prediction variance, which quantifies the model’s temporal sensitivity to different frame regions. Only QA pairs with variance above an empirical threshold $\tau {=} 0.2$ are retained, ensuring that the selected samples exhibit non-trivial temporal diversity. This filtering process yields a filtered dataset of $393$K QA pairs, which we denote as \textit{ReFoCUS}-393K. The curated pool provides more discriminative and temporally grounded feedback for policy optimization, promoting consistent learning dynamics throughout training.

\subsection{Learning Policy under Reward Guidance}
\label{sec:modeling}
\paragraph{Query-Conditioned Policy Architecture.}
To semantically align visual cues in videos with user queries,
the policy model needs to jointly interpret the intent of the query and identify the visual evidence necessary for answering it.
Rather than relying on a predefined set of fixed frames, our approach aims to explore the combinatorial frame space and optimize the frame selection based on contextual relevance.
To achieve this, the policy architecture must satisfy two essential properties:
(\lowercase\expandafter{\romannumeral1}) handling dense temporal visual inputs efficiently while capturing long-range dependencies, and
(\lowercase\expandafter{\romannumeral2}) selecting frames in an \textit{autoregressive} manner conditioned on user queries and previously chosen frames, enabling sequential exploration of the frame space, since explicit teacher-forcing supervision at the frame-level is not available during the training.

Accordingly, we implement the policy model using Video-MA$^2$mba~\cite{lee2024look}, which ensures linear computational and memory complexity through its state-space sequence modeling~\cite{Dao2024TransformersAS}, while preserving temporal dependencies across long video sequences. Leveraging its partially Markov property further reduces memory usage and preserves causal consistency, enabling each frame-selection step to depend only on the query $\langle v, t\rangle$ and the previously selected frames $f_{<i}$, thereby maintaining temporal coherence without redundant conditioning.

Since we adopt Video-MA$^2$mba, which is SFT-pretrained with video-language instruction tuning, it already possesses fundamental visual and linguistic reasoning abilities. We repurpose these pretrained perceptual representations into frame-selection capabilities by removing the final unembedding head, reinitializing the terminal layers, and fine-tuning it to autoregressively predict the next frame conditioned on the query and prior selections. As illustrated in~\cref{fig:3}, the policy starts from a special token \texttt{<sof>} (start-of-frame) and sequentially generates query embeddings $q$ that attend over the pool of frame embeddings $k$ to produce a softmax-normalized probability distribution. The next frame is then sampled according to $f_i {\sim} \pi_\theta(\cdot \mid f_{<i}, v, t)$ until $T'$ unique frames are selected, followed by temporal sorting and forwarding to the video-LLMs for reasoning.

\vspace{-4mm}
\paragraph{Reward Modeling.}
Having established the autoregressive policy architecture for query-conditioned frame selection, the policy is then optimized to align its selection behavior with the visual utility of the downstream video-LLMs. Here, we note that a single trajectory from the policy corresponds to one candidate frame subset, but relying on individual samples can potentially yield sparse and noisy feedback. To stabilize optimization, we evaluate sampled subsets in a group-wise manner, allowing the policy to learn from the relative ranking among candidates rather than from absolute reward magnitudes. In this setup, the policy $\pi_\theta$ samples $N$ candidate subsets from each $\langle v,t\rangle$ pair, where each $j$-th subset is denoted as $s_{j}$. Each frame subset generated by $\pi_\theta$ is treated as an individual action whose value reflects how effectively the selected frames enable the reward model $r_\varphi$ to answer the question correctly.

We estimate this value using the model’s predictive distribution over answer options and explicitly maximize the decision margin between the ground-truth answer $y^{*}$ and the most competitive incorrect choice $\tilde{y}$, which captures how confidently the model favors the correct answer given the selected frames. The margin-based reward is defined as the prediction difference:
\begin{equation}
\label{eq:reward}
r_j = \frac{r_\varphi(y^{*} \mid s_j, v, t) - r_\varphi(\tilde{y} \mid s_j, v, t)}{r_\varphi(y^{*} \mid s_j, v, t) + r_\varphi(\tilde{y} \mid s_j, v, t)},
\end{equation}
where $r_j$ is normalized within $[-1,1]$ before being used in policy optimization. Our reward design captures both positive and negative evidence by measuring the relative confidence between the correct and distractor answers, and is utilized in a group-wise manner to provide stable and query-conditioned signals for subsequent policy optimization.

\paragraph{Frame-level Policy Optimization.}
Building upon the group-relative prediction margin reward above, we optimize the policy $\pi_{\theta}$ through an online policy-gradient objective, maximizing the expected margin-based reward estimated from $r_{\varphi}$. Each sampled subset $s_j$ represents a complete selection trajectory, treated as an atomic action within the policy space. Leveraging these group-wise rewards, we apply GRPO~\cite{shao2024deepseekmath}, an online policy-gradient method that eliminates the need for an explicit value function by using group-normalized rewards as implicit baselines. The normalized advantage for each frame-level trajectory within the $\langle v, t\rangle$ group is computed as:
\begin{equation}
\label{eq:advantage}
    \hat{A}_{j} = \dfrac{r_{j} - \mathrm{mean}(\{r_j\}_{j=1}^{N})}{\mathrm{std}(\{r_j\}_{j=1}^{N})}, 
\end{equation}
where it serves as a normalized learning signal for updating the policy parameters, guiding frame-level selection toward trajectories with higher relative rewards while preserving temporal diversity.

The policy $\pi_{\theta}$ is optimized through a modified GRPO objective for our frame-level optimization setting. Since no pretrained reference policy exists to provide frame-level supervision, we adapt the standard GRPO formulation used in LLM preference optimization~\cite{shao2024deepseekmath, yu2025dapo} by replacing the KL regularization term with an entropy bonus $\mathcal{H}(\pi_\theta)$.
The resulting objective encourages diverse yet reward-guided exploration over the frame-selection space, allowing the policy to progressively align its selection behavior with the confidence distribution of the reward model:
\begin{align}
\label{eq:grpo}
\mathcal{J}(\theta)=
\mathbb{E}_{s_j \sim \pi_{\text{old}}}
\!\left[
\frac{1}{N}\!\sum_{j=1}^{N}
\frac{\pi_\theta(s_j)}{\pi_{\text{old}}(s_j)}\hat{A}_j
+ \beta \mathcal{H}(\pi_\theta)
\right],
\end{align}
where $\mathcal{H}(\pi_\theta)$ acts as a soft regularizer that serves as a practical surrogate for the KL constraint in standard GRPO. It stabilizes policy updates and prevents degenerate frame-selection behaviors such as densely choosing early frames or repeatedly focusing on redundant temporal regions, while maintaining balanced coverage across the video. This reward-guided online policy-gradient optimization enables efficient frame-level learning without requiring a handcrafted or supervised reference policy.

\paragraph{Search-Space Scaling Curriculum.}
Although the proposed policy optimization effectively aligns frame selection with reward feedback, the combinatorial search space remains prohibitively large (\textit{e.g.,} identifying $32$ informative frames out of $512$ candidates involves approximately $\binom{512}{32}\!\approx\!7\!\times\!10^{50}$ possible combinations). Such an enormous space makes it difficult for the policy to consistently locate informative subsets, often leading to unstable gradients or diluted rewards when frame selections are sparse or misaligned. To alleviate it, we introduce a search-space scaling curriculum that progressively expands the exploration scope, enabling the policy to first stabilize to simpler configurations before handling the full selection space.

For example, selecting 4 frames instead of 32 reduces the search space to $\binom{512}{4}\!\approx\!2.8\!\times\!10^{9}$, allowing the policy to explore more intensively and obtain meaningful reward signals without being overwhelmed by search sparsity. This early-stage constraint provides clearer guidance and facilitates stable convergence, as the policy learns which frames are most critical for answering queries. As training progresses, we gradually increase the selection steps $T'$ as $4\!\rightarrow\!8\!\rightarrow\!16\!\rightarrow\!32$, expanding the search space in stages while maintaining reward consistency. Empirically, this curriculum accelerates early learning and mitigates the sparsity-induced instability often observed in large-scale frame-selection policies.
\definecolor{positive}{RGB}{0,0,0}
\definecolor{negative}{RGB}{0,0,0}

\newcommand{\gain}[1]{%
  \ifthenelse{\lengthtest{#1 pt > 0pt}}
    {\textcolor{positive}{\scriptsize$\uparrow$#1}}
    {\textcolor{negative}{\scriptsize$\downarrow$#1}}
}

\begin{table*}[t]

\caption{
Comparison between uniform sampling and \textit{ReFoCUS} across Video-LLMs under a 32-frame budget.
We evaluate open- and closed-source video-LLMs using the \textit{ReFoCUS} frame-selection policy and compare them with uniform sampling on \textit{Video-MME}~\cite{fu2024video}, \textit{LongVideoBench}~\cite{wu2024longvideobench}, \textit{MLVU}~\cite{zhou2024mlvu}, \textit{Video-MMMU}~\cite{hu2025video}, and \textit{NExT-QA} (open-ended)~\cite{xiao2021next}.
}
\label{tab:main_table}
\centering
\small
\resizebox{0.95\linewidth}{!}{
\begin{tabular}{@{}l c l l l l l l l l @{}}
\Xhline{2\arrayrulewidth}
\multicolumn{2}{c}{} & \multicolumn{4}{c}{\textbf{Video-MME (w/o sub)}} & \multicolumn{1}{c}{\textbf{LVB}} & \multicolumn{1}{c}{\textbf{MLVU}} & \multicolumn{1}{c}{\textbf{VMMMU}} & \multicolumn{1}{c}{\textbf{NExT-QA}} \\
\cmidrule(lr){3-6}\cmidrule(lr){7-7}\cmidrule(lr){8-8}\cmidrule(lr){9-9}\cmidrule(lr){10-10}
\textbf{Model} & \textbf{LLM Size} & \multicolumn{1}{c}{short} &  \multicolumn{1}{c}{medium} & \multicolumn{1}{c}{long} & \multicolumn{1}{c}{overall} & \multicolumn{1}{c}{acc. (val)} & \multicolumn{1}{c}{m-avg} & \multicolumn{1}{c}{overall} & \multicolumn{1}{c}{wups (val)} \\
\hline
\rowcolor[HTML]{ECF4FF} \multicolumn{10}{l}{\textit{Closed Source}} \\
Gemini 2.5 Flash~\cite{comanici2025gemini} & \multirow{2}{*}{-} & \textcolor{gray}{77.6} & \textcolor{gray}{63.7} & \textcolor{gray}{56.8} & \textcolor{gray}{66.0} & \textcolor{gray}{47.9} & \textcolor{gray}{52.8} & \textcolor{gray}{40.6} & \textcolor{gray}{11.7} \\
\hspace{0.5em}+ \textbf{ReFoCUS} &  & 79.3 \gain{1.7} & 68.3 \gain{4.6} & 60.8 \gain{4.0} & 69.5 \gain{3.5} & 50.9 \gain{3.0} & 58.0 \gain{5.2} & 45.6 \gain{5.0} & 11.9 \gain{0.2} \\
GPT-4o~\cite{gpt4o} & \multirow{2}{*}{-} & \textcolor{gray}{68.0} & \textcolor{gray}{55.0} & \textcolor{gray}{53.3} & \textcolor{gray}{58.8} & \textcolor{gray}{49.5} & \textcolor{gray}{58.7} & \textcolor{gray}{62.9} & \textcolor{gray}{8.5} \\
\hspace{0.5em}+ \textbf{ReFoCUS} &  & 68.2 \gain{0.2} & 60.1 \gain{5.1} & 54.0 \gain{0.7} & 60.8 \gain{2.0} & 52.9 \gain{3.4} & 65.1 \gain{6.4} & 62.1 \gain{-0.8} & 9.1 \gain{0.6} \\
\hline
\rowcolor[HTML]{ECF4FF} \multicolumn{10}{l}{\textit{Open Source}} \\

LLaVA-OneVision~\cite{li2024llava} & \multirow{2}{*}{0.5B} & \textcolor{gray}{53.7} & \textcolor{gray}{39.9} & \textcolor{gray}{37.0} & \textcolor{gray}{43.5} & \textcolor{gray}{44.7} & \textcolor{gray}{44.8} & \textcolor{gray}{17.3} & \textcolor{gray}{18.1} \\
\hspace{0.5em}+ \textbf{ReFoCUS} &  & 58.3 \gain{4.6} & 44.6 \gain{4.7} & 38.3 \gain{1.3} & 47.1 \gain{3.6} & 48.7 \gain{4.0} & 50.3 \gain{5.5} & 19.4 \gain{2.1} & 18.9 \gain{0.8} \\

InternVL3~\cite{zhu2025internvl3} & \multirow{2}{*}{1B} & \textcolor{gray}{63.1} & \textcolor{gray}{46.9} & \textcolor{gray}{39.9} & \textcolor{gray}{50.0} & \textcolor{gray}{47.6} & \textcolor{gray}{54.0} & \textcolor{gray}{27.7} & \textcolor{gray}{20.0} \\
\hspace{0.5em}+ \textbf{ReFoCUS} &  & 66.4 \gain{3.3} & 51.8 \gain{4.9} & 42.6 \gain{2.7} & 53.6 \gain{3.6} & 50.6 \gain{3.0} & 58.9 \gain{4.9} & 29.3 \gain{1.6} & 20.4 \gain{0.4} \\

\cdashline{1-10}\noalign{\vskip 0.2ex}

VideoLLaMA 3~\cite{zhang2025videollama} & \multirow{2}{*}{2B} & \textcolor{gray}{55.2} & \textcolor{gray}{38.8} & \textcolor{gray}{35.2} & \textcolor{gray}{43.1} & \textcolor{gray}{48.8} & \textcolor{gray}{46.8} & \textcolor{gray}{28.7} & \textcolor{gray}{18.9} \\
\hspace{0.5em}+ \textbf{ReFoCUS} &  & 58.9 \gain{3.7} & 44.1 \gain{5.3} & 38.3 \gain{3.1} & 47.1 \gain{4.0} & 53.7 \gain{4.9} & 50.2 \gain{3.4} & 29.2 \gain{0.5} & 20.7 \gain{1.8} \\

InternVL3~\cite{zhu2025internvl3} & \multirow{2}{*}{2B} & \textcolor{gray}{71.0} & \textcolor{gray}{56.4} & \textcolor{gray}{47.8} & \textcolor{gray}{58.4} & \textcolor{gray}{50.9} & \textcolor{gray}{62.7} & \textcolor{gray}{38.3} & \textcolor{gray}{24.4} \\
\hspace{0.5em}+ \textbf{ReFoCUS} &  & 72.2 \gain{1.2} & 60.2 \gain{3.8} & 49.7 \gain{1.9} & 60.7 \gain{2.3} & 54.9 \gain{4.0} & 68.0 \gain{5.3} & 39.3 \gain{1.0} & 25.0 \gain{0.6} \\

InternVL3.5~\cite{wang2025internvl3} & \multirow{2}{*}{4B} & \textcolor{gray}{76.4} & \textcolor{gray}{60.3} & \textcolor{gray}{51.3} & \textcolor{gray}{62.7} & \textcolor{gray}{57.7} & \textcolor{gray}{66.6} & \textcolor{gray}{52.0} & \textcolor{gray}{22.1} \\
\hspace{0.5em}+ \textbf{ReFoCUS} &  & 78.0 \gain{1.6} & 62.3 \gain{2.0} & 57.4 \gain{6.1} & 65.9 \gain{3.2} & 62.6 \gain{4.9} & 71.5 \gain{4.9} & 53.3 \gain{1.3} & 22.9 \gain{0.8} \\

Qwen3-VL~\cite{yang2025qwen3} & \multirow{2}{*}{4B} & \textcolor{gray}{74.1} & \textcolor{gray}{61.0} & \textcolor{gray}{51.3} & \textcolor{gray}{62.1} & \textcolor{gray}{57.4} & \textcolor{gray}{63.1} & \textcolor{gray}{54.0} & \textcolor{gray}{23.8} \\
\hspace{0.5em}+ \textbf{ReFoCUS} &  & 76.7 \gain{2.6} & 65.7 \gain{4.7} & 57.0 \gain{5.7} & 66.4 \gain{4.3} & 61.9 \gain{4.5} & 71.9 \gain{8.8} & 56.4 \gain{2.4} & 24.1 \gain{0.3} \\

\cdashline{1-10}\noalign{\vskip 0.2ex}

VideoLLaMA 3~\cite{zhang2025videollama} & \multirow{2}{*}{7B} & \textcolor{gray}{70.4} & \textcolor{gray}{57.7} & \textcolor{gray}{48.9} & \textcolor{gray}{59.0} & \textcolor{gray}{54.8} & \textcolor{gray}{52.9} & \textcolor{gray}{32.8} & \textcolor{gray}{25.8} \\
\hspace{0.5em}+ \textbf{ReFoCUS} &  & 72.2 \gain{1.8} & 60.1 \gain{2.4} & 54.3 \gain{5.4} & 62.2 \gain{3.2} & 57.0 \gain{2.2} & 59.8 \gain{6.9} & 34.4 \gain{1.6} & 26.5 \gain{0.7} \\

LLaVA-OneVision~\cite{li2024llava} & \multirow{2}{*}{7B} & \textcolor{gray}{70.9} & \textcolor{gray}{55.7} & \textcolor{gray}{48.8} & \textcolor{gray}{58.4} & \textcolor{gray}{55.0} & \textcolor{gray}{63.7} & \textcolor{gray}{34.1} & \textcolor{gray}{16.2} \\
\hspace{0.5em}+ \textbf{ReFoCUS} &  & 72.8 \gain{1.9} & 61.7 \gain{6.0} & 53.4 \gain{4.6} & 62.6 \gain{4.2} & 61.0 \gain{6.0} & 68.5 \gain{4.8} & 35.7 \gain{1.6} & 16.4 \gain{0.2} \\

InternVL3~\cite{zhu2025internvl3} & \multirow{2}{*}{8B} & \textcolor{gray}{75.1} & \textcolor{gray}{64.4} & \textcolor{gray}{53.4} & \textcolor{gray}{64.3} & \textcolor{gray}{57.8} & \textcolor{gray}{68.1} & \textcolor{gray}{49.3} & \textcolor{gray}{26.6} \\
\hspace{0.5em}+ \textbf{ReFoCUS} &  & 75.8 \gain{0.7} & 66.8 \gain{2.4} & 58.3 \gain{4.9} & 67.0 \gain{2.7} & 62.0 \gain{4.2} & 72.7 \gain{4.6} & 50.6 \gain{1.3} & 26.8 \gain{0.2} \\

InternVL3.5~\cite{wang2025internvl3} & \multirow{2}{*}{8B} & \textcolor{gray}{77.4} & \textcolor{gray}{62.4} & \textcolor{gray}{53.2} & \textcolor{gray}{64.4} & \textcolor{gray}{59.7} & \textcolor{gray}{67.3} & \textcolor{gray}{50.0} & \textcolor{gray}{24.3} \\
\hspace{0.5em}+ \textbf{ReFoCUS} &  & 76.2 \gain{-1.2} & 64.9 \gain{2.5} & 58.9 \gain{5.7} & 66.7 \gain{2.3} & 64.1 \gain{4.4} & 70.6 \gain{3.3} & 53.2 \gain{3.2} & 24.7 \gain{0.4} \\

Qwen3-VL~\cite{yang2025qwen3} & \multirow{2}{*}{8B} & \textcolor{gray}{75.1} & \textcolor{gray}{64.6} & \textcolor{gray}{55.3} & \textcolor{gray}{65.0} & \textcolor{gray}{56.6} & \textcolor{gray}{63.0} & \textcolor{gray}{59.1} & \textcolor{gray}{25.3} \\
\hspace{0.5em}+ \textbf{ReFoCUS} &  & 79.6 \gain{4.5} & 67.0 \gain{2.4} & 58.9 \gain{3.6} & 68.5 \gain{3.5} & 63.3 \gain{6.7} & 72.5 \gain{9.5} & 61.1 \gain{2.0} & 25.7 \gain{0.4} \\
\Xhline{2\arrayrulewidth}
\end{tabular}
}
\vspace{-3mm}
\end{table*}


\section{Experiments}
\paragraph{Implementation \& Training Details.}
As aforementioned, we adopt Video-MA$^2$mba~\cite{lee2024look} as the backbone of the policy model $\pi_{\theta}$, while the last 18 Mamba blocks are re-initialized to enhance task-specific flexibility. We further add new unembedding layers for frame and query embeddings. Frames are sampled at $4\,\mathrm{fps}$ up to 512 frames, beyond which uniform subsampling is applied. Starting from the special token \texttt{<sof>}, the model autoregressively generates $T'$ frame tokens. We employ InternVL3~\cite{zhu2025internvl3} as the reward model $r_{\varphi}$ to provide reliable reward feedback for policy optimization. During training, $T'$ is progressively increased across curriculum stages $(4{\rightarrow}8{\rightarrow}16{\rightarrow}32)$, which exponentially enlarges the frame-selection search space. To ensure stable optimization under this expansion, both the learning rate and entropy coefficient $\beta$ are gradually reduced as the curriculum advances. Training is performed in \textit{bfloat16} precision using the AdamW~\cite{loshchilov2017decoupled} optimizer with a linear warm-up and cosine decay scheduler, requiring approximately $1$k H200 GPU hours. Detailed stage-wise hyperparameter configurations are provided in~\cref{sec:supp_training}.

\vspace{-4mm}
\paragraph{Benchmarks \& Baselines.} We evaluate our method across five representative video understanding benchmarks, which span a wide evaluation video QA spectrum. For fair comparison, we report mean scores over four fixed random seeds. Video-MME~\cite{fu2024video} provides assessments for LMMs across short-to-long videos. LongVideoBench~\cite{wu2024longvideobench} focuses on evaluating temporal localization and long-context comprehension through multiple-choice QA. MLVU~\cite{zhou2024mlvu} expands the evaluation space with multi-task and multi-granular assessments (\textit{e.g.,} global summarization and fine-grained temporal reasoning). Video-MMMU~\cite{hu2025video} emphasizes knowledge acquisition across perception, comprehension, and adaptation, using professional educational videos to evaluate a model's ability to learn new information. NExT-QA~\cite{xiao2021next} benchmarks open-ended video question answering, emphasizing causal and temporal reasoning over natural video scenes. We build \textit{ReFoCUS} upon a diverse set of open-~\cite{li2024llava, zhang2025videollama, bai2025qwen2, zhu2025internvl3, wang2025internvl3, yang2025qwen3} and closed-source~\cite{comanici2025gemini, gpt4o} models with different model scales.

\subsection{Main Results of Video Understanding}
\label{sec:main_result}
\paragraph{Performance on Task-diverse Video QA Benchmarks.} 
As shown in~\cref{tab:main_table}, integrating \textit{ReFoCUS} consistently improves performance across all video QA benchmarks and various model scales, including both open-/closed-source models, which suggests our policy model effectively gathers query-conditioned visual evidence that is broadly useful for downstream reasoning. The learned policy compresses sparsely distributed temporal cues into a concise yet informative set of frames, yielding consistent improvements even in open-ended NExT-QA without answer options and demonstrating its transferability beyond the multiple-choice format used in training. Moreover, even for publicly available models pretrained with uniform or fixed-interval frame sampling, the information contained in \textit{ReFoCUS}-selected frames proves to be more decisive than potential disturbances from irregular time stamps. Overall, our findings demonstrate that even in the absence of explicit frame-level supervision, reinforcement learning alone, when implemented with \textit{ReFoCUS}, is sufficient to assemble the most query-relevant visual cues for comprehensive reasoning.

\begin{figure}[t]
\centering
\includegraphics[width=0.99\linewidth]{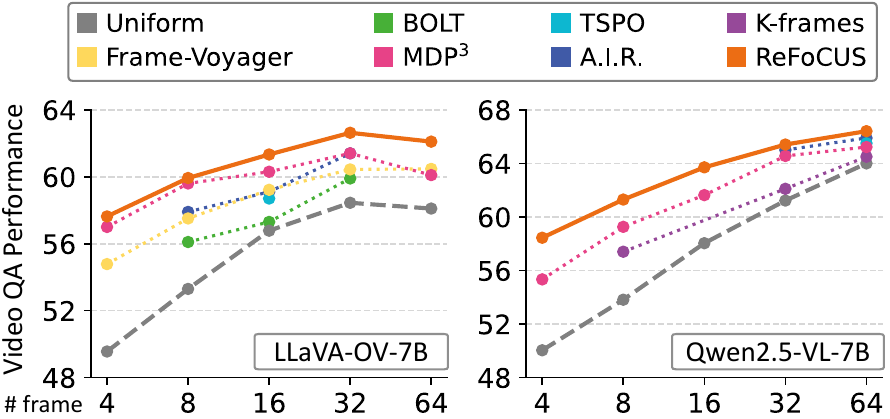}
\vspace{-5mm}
\caption{Performance comparison with other frame selection baselines 
on Video-MME across 4–64 frames.}
\vspace{-3mm}
\label{fig:4}
\end{figure}


\begin{table}[t]
\caption{Efficiency comparison between baselines and frame selection methods with 512-frame input.}
\label{tab:computational_metrics}
\centering
\small
\resizebox{1.0\linewidth}{!}{
\begin{tabular}{lcccc}
\Xhline{2\arrayrulewidth}
\textbf{Model} & \textbf{Size} & \textbf{TFLOPs} & \textbf{Time}(s) & \textbf{Mem}(GB) \\
\hline
\rowcolor[HTML]{ECF4FF} \multicolumn{5}{l}{\textit{Downstream Baselines}} \\
LLaVA-OV~\cite{li2024llava}      & 7B & 1780 & 62 & 77.3 \\
InternVL3~\cite{zhu2025internvl3}      & 8B  & 2099 & 46 & 44.6 \\
\hline
\rowcolor[HTML]{ECF4FF} \multicolumn{5}{l}{\textit{Frame Selection Methods (added to baselines)}} \\
+ VideoAgent~\cite{wang2024videoagent}    & GPT-4 & 691 & -   & -    \\
+ \textsc{Frame-Voyager}~\cite{yu2025framevoyager} & -    & -   & -   & 23.0    \\
+ \textit{T}*~\cite{ye2025re}            & 8B  & 393 & 35   & 19.1    \\
+ \textsc{mDP$^3$}~\cite{sun2025mdp3}          & 0.9B & 309 & 10  & 17.7 \\
+ \textbf{ReFoCUS}  & 1.8B & 428 & 9  & 5.3  \\
\Xhline{2\arrayrulewidth}
\end{tabular}
}
\vspace{-5.0mm}
\end{table}


\paragraph{Comparison with Other Frame Selection Baselines.}
To demonstrate the advantages of \textit{ReFoCUS}, we conduct a comprehensive comparison against existing frame-selection methods across two downstream video-LLMs~\cite{li2024llava,bai2025qwen2} on Video-MME benchmark, varying the frame budget from $4$ to $64$. As shown in~\cref{fig:4}, \textit{ReFoCUS} consistently surpasses all existing methods across the entire frame-budget range. Compared to both heuristic~\cite{sun2025mdp3,liu2025bolt,zou2025air} and learning-based~\cite{yu2025framevoyager,tang2025tspo,yao2025k} baselines, \textit{ReFoCUS} demonstrates a stronger ability to identify query-conditioned frames that carry highly informative visual cues for answering downstream questions. Furthermore, its steady improvement with increasing frame budgets indicates that the reward-aligned policy learned through reinforcement learning can successfully transfer to general video-LLMs, effectively selecting frames that enhance overall video understanding.

\subsection{Ablation Study for ReFoCUS}
\paragraph{Computational Efficiency Analysis.}
To analyze the relative computational cost of \textit{ReFoCUS}, we measured the processing time, TFLOPs, and peak GPU memory usage of each method when handling 512-frame video inputs under the 32-frame selection setting. For fair comparison, all video frames were preloaded to exclude data-loading overhead, ensuring that the reported time reflects only the core computation of each method. FLOPs were measured based on the actual operations executed within the PyTorch framework, and the average runtime was measured on a single NVIDIA A100 GPU. As summarized in~\cref{tab:computational_metrics}, \textit{ReFoCUS} achieves a strong balance between computational efficiency and model capacity. Despite operating with a moderately sized 1.8B parameter policy, it performs frame selection substantially faster and with lower GPU memory consumption compared to existing baselines such as \textsc{mDP$^3$}~\cite{sun2025mdp3} and \textit{T}*~\cite{ye2025re}. This efficiency gain primarily stems from the use of a lightweight policy model that relies solely on its intrinsic computational capacity, without any iterative evaluation or algorithmic post-processing. By simplifying the frame selection process into a single feed-forward, autoregressive generation step, \textit{ReFoCUS} achieves both computational efficiency and practical scalability.

\providecommand{\fivemodelsfn}{}
\providecommand{\fivemodelsfnEmpty}{}

\begin{table}[t]

\caption{
Effect of policy and reward model configurations, averaged across five\fivemodelsfn\ 7--8B models with 32-frame inputs.
}
\label{tab:policy_scaling_multi_bench}

\centering
\small
\resizebox{1.0\linewidth}{!}{
\begin{tabular}{@{}cccccc@{}}
\Xhline{2\arrayrulewidth}
\multicolumn{2}{c}{\textbf{Train Configuration}} &
\multirow[c]{2}{*}{\textbf{VMME}} &
\multirow[c]{2}{*}{\textbf{LVB}} &
\multirow[c]{2}{*}{\textbf{MLVU}} &
\multirow[c]{2}{*}{\textbf{VMMMU}} \\
\textbf{$\pi_{\theta}$} & \textbf{$r_{\varphi}$} &
    &  &  &  \\
\hline
\rowcolor[HTML]{ECF4FF} \multicolumn{6}{l}{\textit{Effect of Policy Model Scale}} \\
0.3B & InternVL3 2B  & 64.8 & 59.2 & 66.3 & 45.9 \\
1.3B & InternVL3 2B  & 65.4 & 61.4 & 68.8 & 46.5 \\
2.7B & InternVL3 2B  & 64.5 & 60.9 & 70.1 & 46.7 \\
\hline
\rowcolor[HTML]{ECF4FF} \multicolumn{6}{l}{\textit{Effect of Reward Model Choice}} \\
1.3B & InternVL3 2B  & 65.4 & 61.4 & 68.8 & 46.5 \\
1.3B & Qwen3-VL 4B   & 65.0 & 61.2 & 70.9 & 46.2 \\
1.3B & LLaVA-OV 7B   & 64.0 & 61.0 & 68.6 & 45.1 \\
\Xhline{2\arrayrulewidth}
\end{tabular}
}
\vspace{-4mm}
\end{table}
\ifx\fivemodelsfn\fivemodelsfnEmpty\else
\begingroup
\setlength{\skip\footins}{2pt}%
\renewcommand{\footnoterule}{\vspace*{-3pt}\hrule width 0.3\columnwidth height 0.3pt\vspace*{1pt}}%
\footnotetext[1]{\label{fn:five_models}\scriptsize VideoLLaMA3, LLaVA-OneVision, InternVL3, InternVL3.5, and Qwen3VL}%
\endgroup
\fi

\begin{table}[t]

\caption{
Performance by curriculum stages, averaged across five\fivemodelsfn\ 7--8B models and evaluated on Video-MME (w/o sub.).
}
\label{tab:curriculum_performance}

\centering
\small
\resizebox{1.0\linewidth}{!}{
\begin{tabular}{lllll}
\Xhline{2\arrayrulewidth}
\multirow{2}{*}{\textbf{Setting}} & \multicolumn{4}{c}{\textbf{Evaluation Frame Budget}} \\
\cmidrule(lr){2-5}
    & \multicolumn{1}{c}{4 frame} & \multicolumn{1}{c}{8 frame} & \multicolumn{1}{c}{16 frame} & \multicolumn{1}{c}{32 frame} \\
\hline
Uniform & 53.3 & 56.5 & 59.7 & 62.2 \\ \hline
\rowcolor[HTML]{ECF4FF} \multicolumn{5}{l}{\textit{ReFoCUS w/o Search-Space Scaling Curriculum}} \\
32 & 55.3 {\scriptsize +2.0} & 58.5 {\scriptsize +2.0} & 60.7 {\scriptsize +1.0} & 62.8 {\scriptsize +0.6} \\ \hline
\rowcolor[HTML]{ECF4FF} \multicolumn{5}{l}{\textit{ReFoCUS w/ Search-Space Scaling Curriculum}} \\
4 & 59.9 {\scriptsize +6.6} & 61.3 {\scriptsize +4.8} & 62.5 {\scriptsize +2.8} & 63.8 {\scriptsize +1.6} \\
4→8 & \textbf{60.1 {\scriptsize +6.8}} & 62.4 {\scriptsize +5.9} & 63.9 {\scriptsize +4.2} & 64.8 {\scriptsize +2.6} \\
4→8→16 & 60.0 {\scriptsize +6.7} & 62.4 {\scriptsize +5.9} & 64.1 {\scriptsize +4.4} & 65.2 {\scriptsize +3.0} \\
4→8→16→32 & \textbf{60.1 {\scriptsize +6.8}} & \textbf{62.7 {\scriptsize +6.2}} & \textbf{64.2 {\scriptsize +4.5}} & \textbf{65.4 {\scriptsize +3.2}} \\
\Xhline{2\arrayrulewidth}
\end{tabular}
}
\vspace{-5mm}
\end{table}


\vspace{-4mm}
\paragraph{Effects of Model Scale and Reward Choice.}
To systematically analyze the impact of both policy model scale and reward model choice, we conduct ablation studies under various configurations. Specifically, we train policies of different sizes and reward models, and evaluate the resulting frame selections with 32-frame inputs, as described in~\cref{tab:policy_scaling_multi_bench}. We observe that a smaller policy (0.3B) leads to degraded performance, whereas scaling up to 2.7B does not necessarily bring further gains. We attribute this to the relatively simpler frame selection task compared to the fine-grained visual reasoning required for complex question answering, where once the policy learns to capture the essential temporal cues, further increases in capacity offer little additional benefit. Furthermore, variations in the reward model led to only marginal changes in performance, indicating that the underlying visual boundaries used to discriminate informative cues are largely shared across different reward sources. This consistency demonstrates that our method operates robustly across both policy scales and reward models, showing minimal sensitivity to either factor.

\vspace{-2mm}
\paragraph{Ablation on Search Space Expansion.}
To analyze the effectiveness of the search space scaling, we conduct an additional ablation study. As shown in~\cref{tab:curriculum_performance}, compared to the non-curriculum setting with \textit{ReFoCUS} (32-frame only) that explores the full space from the beginning (+0.6), the progressive curriculum that gradually increases the selection range up to 32 frames achieves a larger improvement (+3.2). Moreover, performance consistently improves as the curriculum advances through each stage, reaching the highest score at the final step. This trend is observed not only at the 32-frame level but also under sub-budgets (4, 8, and 16), indicating that gradually scaling the reinforcement learning setup from simple to complex configurations plays a crucial role in effectively aligning the policy model.

\begin{figure}[t]
\centering
\includegraphics[width=1.0\linewidth]{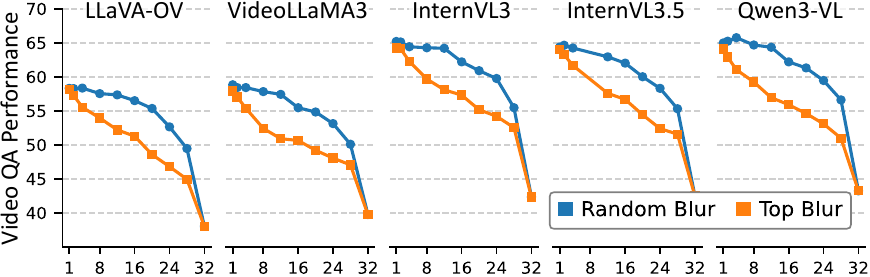}
\vspace{-8mm}
\caption{Performance drop on Video-MME under frame-blur ablation. \textit{Top-blur} progressively corrupts policy-preferred frames, while \textit{Random-blur} blurs random ones.}
\vspace{-3mm}
\label{fig:5}
\end{figure}

\begin{figure}[t]
\centering
\includegraphics[width=1.0\linewidth]{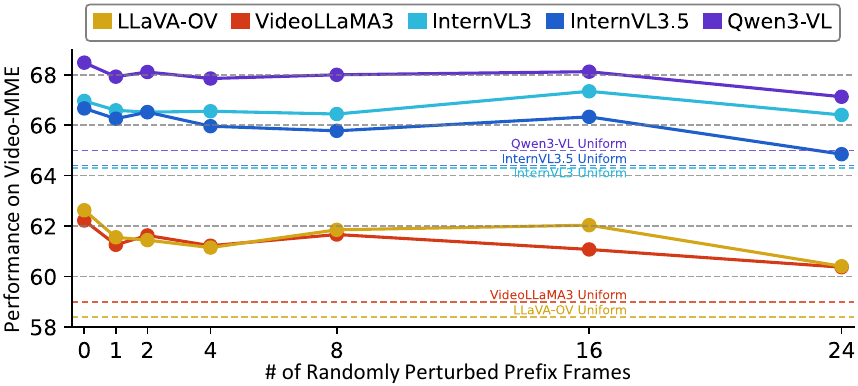}
\vspace{-8mm}
\caption{Early-noise robustness under 32 frame evaluation. The initial selections are replaced with random frames.}
\vspace{-4mm}
\label{fig:6}
\end{figure}

\vspace{-0.5mm}
\subsection{Additional Analysis on ReFoCUS}
\vspace{-0.5mm}
\paragraph{Semantic Grounding of Frame Selections.}
To understand whether the policy's choices genuinely reflect meaningful visual cues, we perform an evidence-suppression ablation that verifies how the video-LLM's reasoning degrades when the frames most frequently sampled by the policy are progressively blurred. To estimate frame importance, we aggregate the policy’s sampling trajectories over 64 runs and compute the cumulative selection frequency of each frame, identifying those most consistently chosen across trajectories.
We then apply Gaussian blur ($\sigma{=}200$px) to the top-$k$ frames (\textit{Top blur}) and compare the degradation against randomly blurred frames (\textit{Random blur}). As observed in~\cref{fig:5}, suppressing these high-evidence frames leads to a pronounced accuracy collapse, showing that \textit{ReFoCUS} consistently focuses on frames that are visually and semantically indispensable.

\paragraph{Robustness to Frame Perturbation.}
Since the \textit{ReFoCUS} policy is trained online and only observes the frame sequences it selects during rollouts, we further examine whether the learned policy remains robust when the initial selection order is perturbed. To do so, we evaluate \textit{ReFoCUS} under a constrained setting where the first $k$ frames are randomly sampled and prefixed in advance, and the policy continues the selection process for the remaining frames. In~\cref{fig:6}, the model maintains consistently higher performance than uniform sampling across all noise levels, regardless of the degree of prefix perturbation. This supports that even when the policy is forced into unlikely or suboptimal early choices, it can still compose an informative frame set by compensating with subsequent selections, demonstrating robustness and adaptability in its selection strategy.

\begin{figure}[t]
\centering
\includegraphics[width=0.95\linewidth]{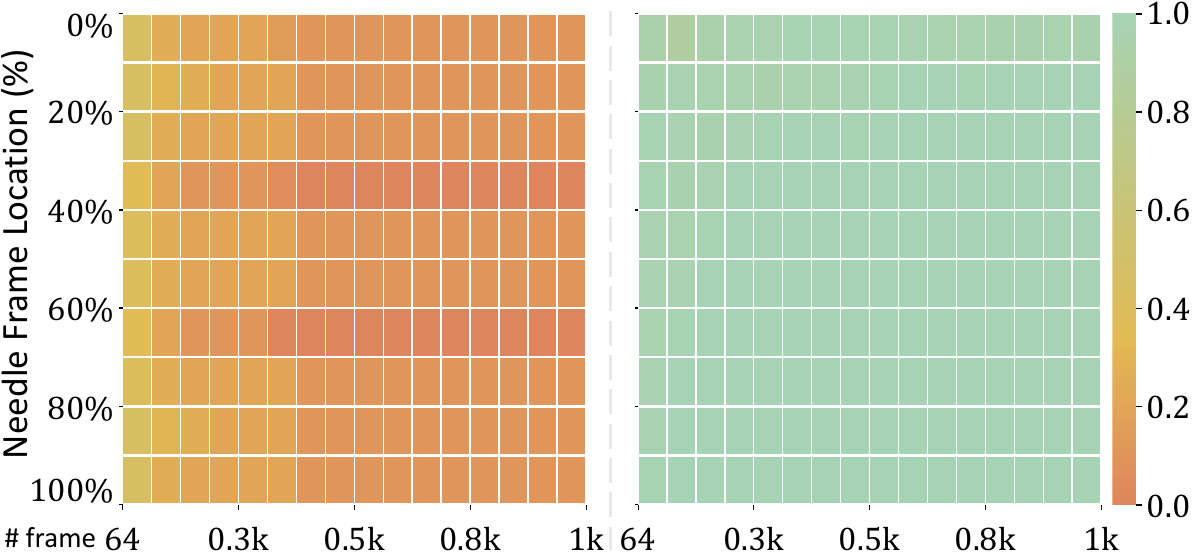}
\vspace*{-3.5mm}
\begin{flushleft}
    \footnotesize
    \hspace{1.4cm}{(a) Uniform Sample \hspace{1.65cm}(b) \textit{ReFoCUS}}
\end{flushleft}
\vspace{-6.7mm}
\caption{
Result of V-NIAH. (a) Uniform sampling (b) Frame selection from \textit{ReFoCUS}. The $x$- and $y$-axis denote \#frames and the needle frame position, respectively.
}
\vspace{-3.2mm}
\label{fig:7}
\end{figure}

\vspace{-1.4mm}
\paragraph{Visual Needle-in-a-Haystack.} To further examine whether \textit{ReFoCUS} can accurately locate task-relevant visual evidence, we conduct a fine-grained analysis under the V-NIAH setup~\cite{zhang2024long}. As illustrated in~\cref{fig:7}(a), the heatmap visualizes uniform sampling using a video-LLM~\cite{zhu2025internvl3}, which fails to capture the temporally sparse but crucial signal (needle frame), as it selects frames uniformly across the entire sequence without regard to content relevance. In contrast, our frame selection (\cref{fig:7}(b)) exhibits a strong concentration on the true needle frame across varying temporal positions, which highlights its ability to precisely localize query-relevant evidence within complex scenes.

\section{Conclusion}
We present \textit{ReFoCUS}, the first framework to apply online policy-gradient reinforcement learning for direct frame-level optimization in video-LLMs. By utilizing frame selection as an autoregressive policy guided by margin-based prediction from reward models, \textit{ReFoCUS} learns to identify semantically rich and temporally relevant frames that align with the model's reasoning trajectory without frame-level supervision. Extensive benchmarks demonstrate consistent gains, validating input-level optimization as a scalable and effective way for advanced multi-modal alignment.

\section{Acknowledgements} This work was supported in part by IITP grant funded by the Korea government (MSIT) (No. RS-2020-II200004, Development of Previsional Intelligence based on Long-Term Visual Memory Network), the Institute of Information \& Communications Technology Planning \& Evaluation (IITP) grant funded by the Korea government (MSIT) (No. RS-2022-II220124), and the supercomputing resources provided by KSC (No. KSC-2025-CRE-0090).

{
    \small
    \bibliographystyle{ieeenat_fullname}
    \bibliography{main}
}

\appendix
\definecolor{qaanswer}{RGB}{57,255,20}
\definecolor{qawrong}{RGB}{214,39,40}

\newcommand{\ANS}[1]{\textcolor{qaanswer}{#1}}
\newcommand{\WA}[1]{\textcolor{qawrong}{#1}}

\newcommand{\MethodNameBox}[1]{\makebox[3.2cm][l]{#1}}

\newcommand{\FrameCell}[3]{%
    \begin{minipage}[t]{0.115\linewidth}
        \centering
        \includegraphics[width=\linewidth]{#1/#2/frame#3.jpg}\\
        \vspace{-6pt}
        {\scriptsize \input{#1/#2/time#3.tex}}
        \vspace{2pt}
    \end{minipage}
}

\newcommand{\RenderMethod}[3]{%

    {\large \textbf{#3}}\\[-17pt]

    \begin{center}
        \includegraphics[width=\linewidth]{#1/#2/timebar.pdf}
    \end{center}
    \vspace{-30pt}

    \begin{center}
    \setlength{\tabcolsep}{0.0025\linewidth}
    \renewcommand{\arraystretch}{1.2}
    \begin{tabular}{c c c c c c c c}
        %
    \begin{minipage}[t]{0.115\linewidth}
        \centering
        \includegraphics[width=\linewidth]{#1/#2/frame1.jpg}\\
        \vspace{-6pt}
        {\scriptsize \input{#1/#2/time1.tex}}
        \vspace{2pt}
    \end{minipage}
 &
        %
    \begin{minipage}[t]{0.115\linewidth}
        \centering
        \includegraphics[width=\linewidth]{#1/#2/frame2.jpg}\\
        \vspace{-6pt}
        {\scriptsize \input{#1/#2/time2.tex}}
        \vspace{2pt}
    \end{minipage}
 &
        %
    \begin{minipage}[t]{0.115\linewidth}
        \centering
        \includegraphics[width=\linewidth]{#1/#2/frame3.jpg}\\
        \vspace{-6pt}
        {\scriptsize \input{#1/#2/time3.tex}}
        \vspace{2pt}
    \end{minipage}
 &
        %
    \begin{minipage}[t]{0.115\linewidth}
        \centering
        \includegraphics[width=\linewidth]{#1/#2/frame4.jpg}\\
        \vspace{-6pt}
        {\scriptsize \input{#1/#2/time4.tex}}
        \vspace{2pt}
    \end{minipage}
 &
        %
    \begin{minipage}[t]{0.115\linewidth}
        \centering
        \includegraphics[width=\linewidth]{#1/#2/frame5.jpg}\\
        \vspace{-6pt}
        {\scriptsize \input{#1/#2/time5.tex}}
        \vspace{2pt}
    \end{minipage}
 &
        %
    \begin{minipage}[t]{0.115\linewidth}
        \centering
        \includegraphics[width=\linewidth]{#1/#2/frame6.jpg}\\
        \vspace{-6pt}
        {\scriptsize \input{#1/#2/time6.tex}}
        \vspace{2pt}
    \end{minipage}
 &
        %
    \begin{minipage}[t]{0.115\linewidth}
        \centering
        \includegraphics[width=\linewidth]{#1/#2/frame7.jpg}\\
        \vspace{-6pt}
        {\scriptsize \input{#1/#2/time7.tex}}
        \vspace{2pt}
    \end{minipage}
 &
        %
    \begin{minipage}[t]{0.115\linewidth}
        \centering
        \includegraphics[width=\linewidth]{#1/#2/frame8.jpg}\\
        \vspace{-6pt}
        {\scriptsize \input{#1/#2/time8.tex}}
        \vspace{2pt}
    \end{minipage}
 \\
        %
    \begin{minipage}[t]{0.115\linewidth}
        \centering
        \includegraphics[width=\linewidth]{#1/#2/frame9.jpg}\\
        \vspace{-6pt}
        {\scriptsize \input{#1/#2/time9.tex}}
        \vspace{2pt}
    \end{minipage}
 &
        %
    \begin{minipage}[t]{0.115\linewidth}
        \centering
        \includegraphics[width=\linewidth]{#1/#2/frame10.jpg}\\
        \vspace{-6pt}
        {\scriptsize \input{#1/#2/time10.tex}}
        \vspace{2pt}
    \end{minipage}
 &
        %
    \begin{minipage}[t]{0.115\linewidth}
        \centering
        \includegraphics[width=\linewidth]{#1/#2/frame11.jpg}\\
        \vspace{-6pt}
        {\scriptsize \input{#1/#2/time11.tex}}
        \vspace{2pt}
    \end{minipage}
 &
        %
    \begin{minipage}[t]{0.115\linewidth}
        \centering
        \includegraphics[width=\linewidth]{#1/#2/frame12.jpg}\\
        \vspace{-6pt}
        {\scriptsize \input{#1/#2/time12.tex}}
        \vspace{2pt}
    \end{minipage}
 &
        %
    \begin{minipage}[t]{0.115\linewidth}
        \centering
        \includegraphics[width=\linewidth]{#1/#2/frame13.jpg}\\
        \vspace{-6pt}
        {\scriptsize \input{#1/#2/time13.tex}}
        \vspace{2pt}
    \end{minipage}
 &
        %
    \begin{minipage}[t]{0.115\linewidth}
        \centering
        \includegraphics[width=\linewidth]{#1/#2/frame14.jpg}\\
        \vspace{-6pt}
        {\scriptsize \input{#1/#2/time14.tex}}
        \vspace{2pt}
    \end{minipage}
 &
        %
    \begin{minipage}[t]{0.115\linewidth}
        \centering
        \includegraphics[width=\linewidth]{#1/#2/frame15.jpg}\\
        \vspace{-6pt}
        {\scriptsize \input{#1/#2/time15.tex}}
        \vspace{2pt}
    \end{minipage}
 &
        %
    \begin{minipage}[t]{0.115\linewidth}
        \centering
        \includegraphics[width=\linewidth]{#1/#2/frame16.jpg}\\
        \vspace{-6pt}
        {\scriptsize \input{#1/#2/time16.tex}}
        \vspace{2pt}
    \end{minipage}
 \\
        %
    \begin{minipage}[t]{0.115\linewidth}
        \centering
        \includegraphics[width=\linewidth]{#1/#2/frame17.jpg}\\
        \vspace{-6pt}
        {\scriptsize \input{#1/#2/time17.tex}}
        \vspace{2pt}
    \end{minipage}
 &
        %
    \begin{minipage}[t]{0.115\linewidth}
        \centering
        \includegraphics[width=\linewidth]{#1/#2/frame18.jpg}\\
        \vspace{-6pt}
        {\scriptsize \input{#1/#2/time18.tex}}
        \vspace{2pt}
    \end{minipage}
 &
        %
    \begin{minipage}[t]{0.115\linewidth}
        \centering
        \includegraphics[width=\linewidth]{#1/#2/frame19.jpg}\\
        \vspace{-6pt}
        {\scriptsize \input{#1/#2/time19.tex}}
        \vspace{2pt}
    \end{minipage}
 &
        %
    \begin{minipage}[t]{0.115\linewidth}
        \centering
        \includegraphics[width=\linewidth]{#1/#2/frame20.jpg}\\
        \vspace{-6pt}
        {\scriptsize \input{#1/#2/time20.tex}}
        \vspace{2pt}
    \end{minipage}
 &
        %
    \begin{minipage}[t]{0.115\linewidth}
        \centering
        \includegraphics[width=\linewidth]{#1/#2/frame21.jpg}\\
        \vspace{-6pt}
        {\scriptsize \input{#1/#2/time21.tex}}
        \vspace{2pt}
    \end{minipage}
 &
        %
    \begin{minipage}[t]{0.115\linewidth}
        \centering
        \includegraphics[width=\linewidth]{#1/#2/frame22.jpg}\\
        \vspace{-6pt}
        {\scriptsize \input{#1/#2/time22.tex}}
        \vspace{2pt}
    \end{minipage}
 &
        %
    \begin{minipage}[t]{0.115\linewidth}
        \centering
        \includegraphics[width=\linewidth]{#1/#2/frame23.jpg}\\
        \vspace{-6pt}
        {\scriptsize \input{#1/#2/time23.tex}}
        \vspace{2pt}
    \end{minipage}
 &
        %
    \begin{minipage}[t]{0.115\linewidth}
        \centering
        \includegraphics[width=\linewidth]{#1/#2/frame24.jpg}\\
        \vspace{-6pt}
        {\scriptsize \input{#1/#2/time24.tex}}
        \vspace{2pt}
    \end{minipage}
 \\
        %
    \begin{minipage}[t]{0.115\linewidth}
        \centering
        \includegraphics[width=\linewidth]{#1/#2/frame25.jpg}\\
        \vspace{-6pt}
        {\scriptsize \input{#1/#2/time25.tex}}
        \vspace{2pt}
    \end{minipage}
 &
        %
    \begin{minipage}[t]{0.115\linewidth}
        \centering
        \includegraphics[width=\linewidth]{#1/#2/frame26.jpg}\\
        \vspace{-6pt}
        {\scriptsize \input{#1/#2/time26.tex}}
        \vspace{2pt}
    \end{minipage}
 &
        %
    \begin{minipage}[t]{0.115\linewidth}
        \centering
        \includegraphics[width=\linewidth]{#1/#2/frame27.jpg}\\
        \vspace{-6pt}
        {\scriptsize \input{#1/#2/time27.tex}}
        \vspace{2pt}
    \end{minipage}
 &
        %
    \begin{minipage}[t]{0.115\linewidth}
        \centering
        \includegraphics[width=\linewidth]{#1/#2/frame28.jpg}\\
        \vspace{-6pt}
        {\scriptsize \input{#1/#2/time28.tex}}
        \vspace{2pt}
    \end{minipage}
 &
        %
    \begin{minipage}[t]{0.115\linewidth}
        \centering
        \includegraphics[width=\linewidth]{#1/#2/frame29.jpg}\\
        \vspace{-6pt}
        {\scriptsize \input{#1/#2/time29.tex}}
        \vspace{2pt}
    \end{minipage}
 &
        %
    \begin{minipage}[t]{0.115\linewidth}
        \centering
        \includegraphics[width=\linewidth]{#1/#2/frame30.jpg}\\
        \vspace{-6pt}
        {\scriptsize \input{#1/#2/time30.tex}}
        \vspace{2pt}
    \end{minipage}
 &
        %
    \begin{minipage}[t]{0.115\linewidth}
        \centering
        \includegraphics[width=\linewidth]{#1/#2/frame31.jpg}\\
        \vspace{-6pt}
        {\scriptsize \input{#1/#2/time31.tex}}
        \vspace{2pt}
    \end{minipage}
 &
        %
    \begin{minipage}[t]{0.115\linewidth}
        \centering
        \includegraphics[width=\linewidth]{#1/#2/frame32.jpg}\\
        \vspace{-6pt}
        {\scriptsize \input{#1/#2/time32.tex}}
        \vspace{2pt}
    \end{minipage}

    \end{tabular}
    \end{center}

    \vspace{-10pt}
}

\newcommand{\RenderAllMethods}[1]{%

    \RenderMethod{#1}{Uniform}{\MethodNameBox{Uniform} Ans: \input{#1/Uniform/response.tex}}
    \vspace{15pt}
    \RenderMethod{#1}{MDP3}{\MethodNameBox{\textsc{mDP$^3$}} Ans: \input{#1/MDP3/response.tex}}
    \vspace{15pt}
    \RenderMethod{#1}{ReFoCUS}{\MethodNameBox{\textit{ReFoCUS}} Ans: \input{#1/ReFoCUS/response.tex}}
}

\newcommand{\RenderQAExample}[1]{%

    \begin{tcolorbox}[colback=gray!10,colframe=black]
    \textbf{Question:}\\
    \input{#1/questions.tex}

    \vspace{6pt}
    \textbf{Options:}
    \begin{itemize}
        \input{#1/options.tex}
    \end{itemize}
    \end{tcolorbox}


    %

    \RenderMethod{#1}{Uniform}{\MethodNameBox{Uniform} Ans: \input{#1/Uniform/response.tex}}
    \vspace{15pt}
    \RenderMethod{#1}{MDP3}{\MethodNameBox{\textsc{mDP$^3$}} Ans: \input{#1/MDP3/response.tex}}
    \vspace{15pt}
    \RenderMethod{#1}{ReFoCUS}{\MethodNameBox{\textit{ReFoCUS}} Ans: \input{#1/ReFoCUS/response.tex}}

}

\newcommand{\RenderQAExampleFigure}[2]{%

    \begin{figure*}[!t]
        \centering
        \begin{minipage}{0.70\linewidth}
            %

    \begin{tcolorbox}[colback=gray!10,colframe=black]
    \textbf{Question:}\\
    \input{#1/questions.tex}

    \vspace{6pt}
    \textbf{Options:}
    \begin{itemize}
        \input{#1/options.tex}
    \end{itemize}
    \end{tcolorbox}


    %

    \RenderMethod{#1}{Uniform}{\MethodNameBox{Uniform} Ans: \input{#1/Uniform/response.tex}}
    \vspace{15pt}
    \RenderMethod{#1}{MDP3}{\MethodNameBox{\textsc{mDP$^3$}} Ans: \input{#1/MDP3/response.tex}}
    \vspace{15pt}
    \RenderMethod{#1}{ReFoCUS}{\MethodNameBox{\textit{ReFoCUS}} Ans: \input{#1/ReFoCUS/response.tex}}

        \end{minipage}
        \caption{#2}
        \label{fig:qaexample-#1}
    \end{figure*}
}

\clearpage
\setcounter{page}{1}
\maketitlesupplementary

\section*{Appendix Contents}

\startcontents[appendices]
\printcontents[appendices]{}{1}{}

\newpage

\section{Details for \textit{ReFoCUS}-393K}
\label{sec:supp_data}
In this section, we elaborate on the data curation pipeline used to construct the \textit{ReFoCUS}-393K dataset. The main manuscript describes our reward-variance filtering framework at a conceptual level, and we probe a pretrained video-LLM with multiple frame subsets and retain only those video-QA pairs that exhibit strong temporal sensitivity. Here, we make this procedure fully concrete by specifying (\lowercase\expandafter{\romannumeral1}) how long videos are decomposed into temporal segments, (\lowercase\expandafter{\romannumeral2}) how each segment is paired with its complementary region, and (\lowercase\expandafter{\romannumeral3}) how these intervals are turned into candidate frame subsets for reward computation. These implementation details precisely characterize the preprocessing that produces the training instances for policy learning.

\subsection{Temporal Window-Complement Construction}
Our goal in constructing frame subsets is to systematically contrast local evidence against the rest of the video, so that temporal sensitivity can be measured in a controlled way. To this end, we first tile each video with overlapping temporal windows that capture localized segments, and then pair each window with its complementary region, which serves as a counterfactual context where that local segment is removed.

Formally, for each video $v$ with total length $T$ frames, we divide it into overlapping temporal segments using a fixed window and stride. Specifically, we use a window size of $w = \lceil T / 8 \rceil$ and a stride of $s = \lceil w / 2 \rceil$, resulting in $8$ temporal windows:
\begin{equation}
\label{eq:temp_window}
W_1{=}[0, w),\ W_2{=}[s, s+w),\ \ldots,\ W_8{=}[0, s) \cup [7s, T).
\end{equation}
The first seven windows behave like a sliding temporal crop that sweeps across the video with $50\%$ overlap, while $W_8$ covers the remaining tail region and wraps back to the beginning if necessary. This construction ensures that long videos are covered in a near-uniform manner, while still providing sufficient overlap to capture events that straddle window boundaries.

\begin{figure}[t!]
\centering
\includegraphics[width=0.99\linewidth]{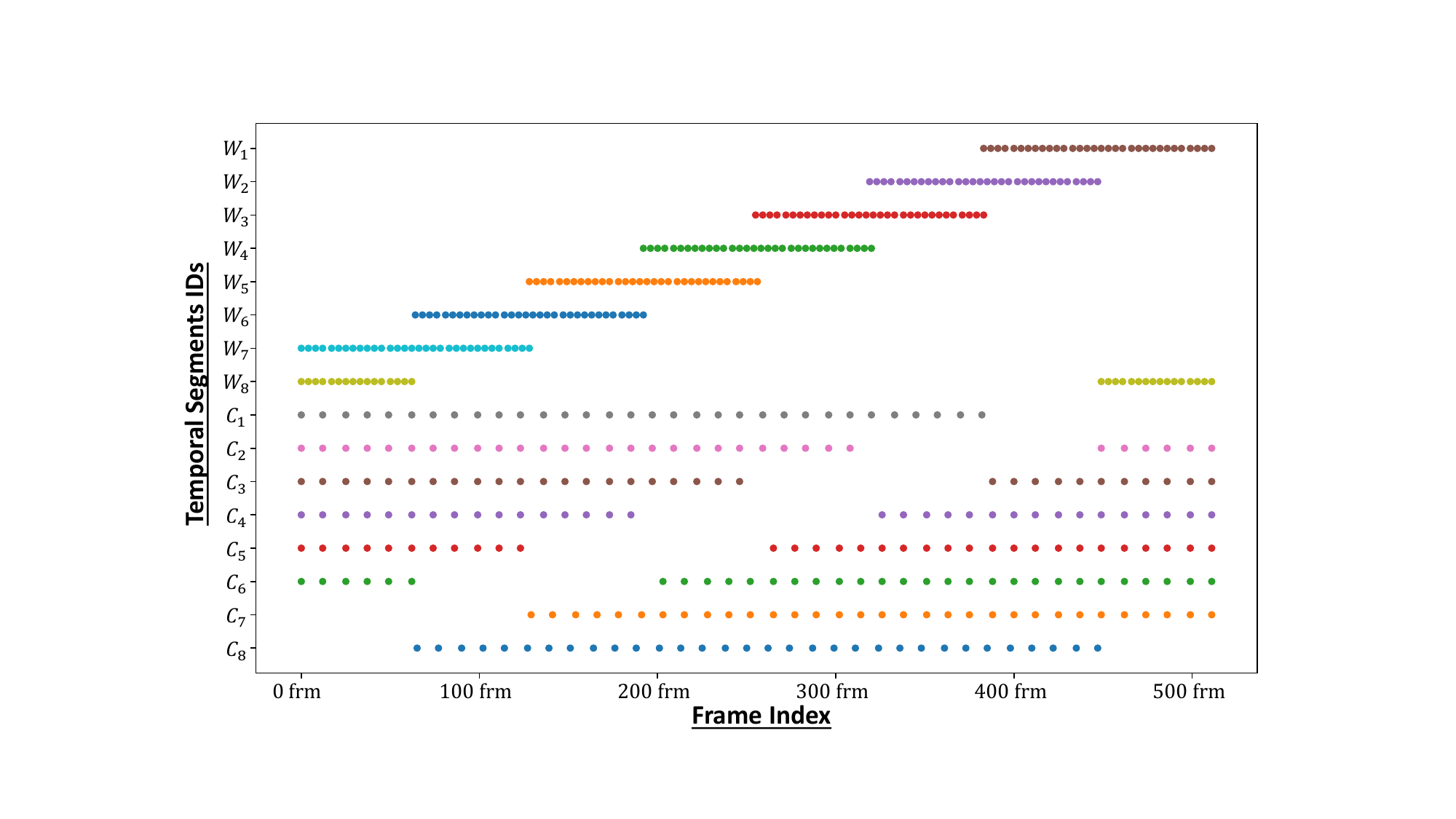}
\vspace{-5mm}
\caption{
Visualization of the temporal segmentation and sampling strategy. We divide each video into $8$ overlapping windows $W_1$ to $W_8$ (top), and for each window, define a complementary region $C_i$ (bottom). We uniformly sample frames from both regions to construct $16$ candidate subsets per QA pair.
}
\label{sfig:1}
\end{figure}

As described in the main manuscript, we do not rely solely on localized segments. Instead, we explicitly construct a complementary region for each window to disentangle the contribution of a short temporal segment from that of the surrounding context. For each window $W_i$, we define its complementary range as $C_i = [0, T) \setminus W_i$, which contains all frames of the video except those in $W_i$. From each of $W_i$ and $C_i$, we then uniformly sample $k = 32$ frames to construct a total of $16$ candidate frame subsets:
\begin{equation}
\label{eq:complementary}
\left\{
\begin{array}{ll}
c_{1}, \ldots, c_{8}  & \text{from windows}, \\
c_{9}, \ldots, c_{16} & \text{from complements.}
\end{array}
\right.
\end{equation}
In other words, each window $W_i$ yields a local subset that focuses on a specific temporal segment and a complementary subset that retains the rest of the video while masking out that segment. By comparing the predictions of the pretrained LMM across these $16$ subsets, we can quantify how much the model’s answer depends on specific temporal regions rather than on global, shortcut-style cues. Figure~\ref{sfig:1} provides a schematic illustration of this sampling scheme, where each $W_i$ (temporal window) and $C_i$ (complement region) is visualized as a distinct dashed line over the frame index axis.

\subsection{Margin Computation over Temporal Regions}
Given the $16$ candidate frame subsets $\{ c_j \}_{j=1}^{16}$ constructed for each video–question pair $\langle v, t \rangle$, we quantify temporal sensitivity using the same margin-based reward employed during policy training. For each subset $c_j$, the frozen reference LMM produces answer logits, from which we compute a prediction margin $r_j$ that reflects how confidently the model favors the correct answer over the strongest distractor. Concretely, $r_j$ is defined as the logit gap between the ground-truth option and the highest-scoring incorrect choice, consistent with the margin computation used in the main training pipeline. The reward variance for a given pair $\langle v, t \rangle$ is then defined as:
\begin{equation}
\label{eq:var_margin}
\mathrm{Var}(m) = \mathrm{Var}\big( \{ r_j \}_{j=1}^{16} \big),
\end{equation}
where the variance measures how strongly the model’s answer confidence changes across different temporal regions of the same video. When the question can be resolved from global context or static cues, the margins $r_j$ remain nearly identical across all subsets, resulting in low variance. Conversely, when the question depends on localized temporal events, informative and uninformative subsets yield markedly different margins, producing higher variance. As a result, $\mathrm{Var}(m)$ provides a direct and interpretable proxy for temporal grounding strength.

\subsection{Variance-based Sample Filtering}
To construct a training pool that emphasizes temporally grounded supervision, we filter each video–QA pair according to its reward variance. For every pair, we compute $\mathrm{Var}(m)$ over the $16$ frame subsets and retain only those whose variance exceeds a threshold of $\tau = 0.2$. This threshold is selected empirically from the overall variance distribution, balancing the removal of flat, low-signal instances with the preservation of a sufficiently large set of temporally informative samples.

Applying this criterion to the full collection of $962$K QA pairs removes examples that exhibit minimal changes in reward across subsets—typically globally answerable questions or instances dominated by dataset-level biases. The resulting filtered dataset contains approximately $393$K QA pairs, which we use as the temporally grounded training pool for subsequent policy optimization.

\subsection{Processing Setup}
All video frames are pre-extracted to maximize inference throughput. For each video–question pair, the 16 frame subsets are generated by indexing into this cached frame pool, and their prediction margins are computed in parallel through a single batched forward pass of the reference LMM. Both raw logits and variance values are recorded for debugging and ablation analyses. This setup ensures stable reward estimation and supports reliable learning dynamics during downstream policy optimization.

\section{Extended Details for \textit{ReFoCUS}}
\label{sec:supp_training}

\subsection{Training Hyperparameters}
During training, for each $\langle v, t\rangle$ instance, the policy generates $N$ frame subsets of length $T'$, which are used for learning. The policy model performs only a single update following each exploration step. We train the policy model based on 1.3B Video-MA$^2$mba and use an InternVL3-2B reward model, with AdamW ($\beta_1{=}0.9$, $\beta_2{=}0.99$), a batch size of 64 $\langle v, t\rangle$ pairs per update, and gradient clipping of 1.0. The learning rate is scheduled by linear warmup of 256 steps followed by cosine decay. Training is conducted in \textit{bfloat16} precision under DeepSpeed ZeRO-1, with input frames resized to $384{\times}384$. Hyperparameters that vary under the Search-Space Scaling Curriculum are summarized in \cref{tab:train_hparams_stage}.

\begin{table}[ht]
\centering
\caption{Stage varying hyperparameters.}
\label{tab:train_hparams_stage}
\small
\vspace{1mm}
\resizebox{\linewidth}{!}{
\begin{tabular}{@{}lccccc@{}}
\toprule
\textbf{Stage} & $T'$ & $N$ & $\beta$ & $\text{LR}_{\text{initialized}}$ & $\text{LR}_{\text{pretrained}}$ \\
\midrule
Stage 1 & 4  & 256 & $1{\times}10^{-3}$  & $6{\times}10^{-5}$ & $1{\times}10^{-5}$ \\
Stage 2 & 8  & 128 & $3{\times}10^{-4}$ & $3{\times}10^{-5}$ & $1{\times}10^{-5}$ \\
Stage 3 & 16 & 64  & $2{\times}10^{-4}$ & $2{\times}10^{-5}$ & $1{\times}10^{-5}$ \\
Stage 4 & 32 & 32  & $1{\times}10^{-4}$ & $2{\times}10^{-5}$ & $1{\times}10^{-5}$ \\
\bottomrule
\end{tabular}
}
\end{table}
\subsection{Frame Selection Process}

Here, we provide the full implementation details of the autoregressive frame-selection mechanism summarized in~\cref{sec:modeling}. The model encodes the input video $v$ and query $t$ to produce frame embeddings $F$, each taken from the final output embedding of its corresponding visual sequence. The backbone's hidden representations ($z_i$ and $F$) are transformed into query $Q$, key $K$, and value $V$ through linear projection followed by RMS-normalization. Starting from \texttt{<sof>}, a query embedding $q_i$ derived from the previously selected frame interacts with all frame keys $K$ to produce the selection score $\mathbf{A}$. These scores define a probability distribution over candidate frames, from which a new frame index $f_i$ is sampled without replacement. The embedding of the selected frame, $V[f_i]$, is then fed back as the selected embedding for the next selection, and this process continues until $T'$ frames are chosen. The detailed autoregressive selection procedure is illustrated in~\cref{alg:arsampling}.

\begin{algorithm}[H]
\caption{\textit{ReFoCUS} Selection}
\label{alg:arsampling}
\begin{algorithmic}[1]
\Require video $v$, user query $t$, selection length $T'$
\State $F \leftarrow \mathrm{Backbone}(v,t)$ \Comment{get frame embedding}
\State $K \leftarrow \mathrm{norm}(F\, W_k^\top)$ \Comment{for key  }
\State $V \leftarrow \mathrm{norm}(F\, W_v^\top)$  \Comment{for value}
\State $\mathbf{e} \leftarrow \texttt{<sof>}$
\For{$i = 1$ to $T'$}
    \State $\mathbf{z}_i \leftarrow \mathrm{BackboneStep}(\mathbf{e})$
    \State $q_i \leftarrow \mathrm{norm}(\mathbf{z}_i W_q^\top)$
    \State $\mathbf{A} = \frac{q_i\, K^\top}{\sqrt{d_{model}}}$  \Comment{get selection score}
    \State $\mathbf{A}[f_{<i}] \leftarrow -\infty$  \Comment{prevent duplicate selection}
    \State $f_i \sim \mathrm{softmax}(\mathbf{A})$ \Comment{equivalent to $\pi_\theta(\cdot \mid f_{<i}, v, t)$}
    \State $\mathbf{e} \leftarrow V[f_i]$ \Comment{get selected embedding}
\EndFor
\State \Return $\{f_i\}_{i=1}^{T'}$
\end{algorithmic}
\end{algorithm}

\subsection{Simplifying the Prediction Margin Reward}

To derive the logit-based, numerically stable simplification of the margin reward from~\cref{eq:reward} introduced in~\cref{sec:modeling}, we expand the reward definition as follows. Each frame subset $s_j$ generated by the policy $\pi_\theta$ is evaluated by the reward model $r_\varphi$, which outputs a confidence distribution over answer candidates. We identify the most competitive incorrect choice $\tilde{y}$ as:
\begin{equation}
\tilde{y} = \arg\max_{y \ne y^{*}} r_\varphi(y \mid s_j, v, t),
\end{equation}
and compute the prediction margin between the correct answer $y^{*}$ and this hardest negative $\tilde{y}$ as:
\begin{equation}
r_j = \frac{r_\varphi(y^{*} \mid s_j, v, t) - r_\varphi(\tilde{y} \mid s_j, v, t)}{r_\varphi(y^{*} \mid s_j, v, t) + r_\varphi(\tilde{y} \mid s_j, v, t)}.
\end{equation}
Letting $z_j(y)$ denote the pre-softmax logits, this margin can be written more simply as:
\begin{equation}
r_j
 = \frac{e^{z_j(y^{*})} - e^{z_j(\tilde{y})}}{e^{z_j(y^{*})} + e^{z_j(\tilde{y})}}
= \tanh\!\left(\frac{z_j(y^{*}) - z_j(\tilde{y})}{2}\right),
\end{equation}
which avoids explicit probability normalization and yields a more numerically stable reward computation.

\section{Additional Experiments}

\subsection{Evaluation on Extra OE Benchmarks}

To examine whether \textit{ReFoCUS} remains effective for open-ended settings, we report additional results on \textit{ActivityNet-QA}~\cite{caba2015activitynet} and \textit{Video-ChatGPT}~\cite{maaz2023video} in~\cref{tab:oe_table}. As in the table, across 7–8B open-sourced models, \textit{ReFoCUS} consistently improves scores on all open-ended evaluation setups.
Similar to the improvements observed on the open-ended subset of \textit{NExT-QA} in~\cref{tab:main_table}, these results confirm that \textit{ReFoCUS} can generalize beyond multiple-choice training and remains effective in open-ended question answering with high transferability.
\begingroup
\definecolor{rfPositive}{RGB}{0,0,0}
\definecolor{rfNegative}{RGB}{0,0,0}

\newcommand{\rfGain}[1]{%
  \begingroup
  \dimen0=#1pt\relax
  \ifdim \dimen0>0pt
    \textcolor{rfPositive}{\scriptsize$\uparrow$#1}%
  \else
    \textcolor{rfNegative}{\scriptsize$\downarrow$#1}%
  \fi
  \endgroup
}

\let\gain\rfGain

\begin{table}[H]
\caption{
Comparison between Baseline (uniform sampling) and \textit{ReFoCUS} on \textit{ActivityNet-QA} and \textit{Video-ChatGPT}. Both Acc (for multiple-choice) and Score (for open-ended responses) are presented on a 0-100 scale.
}
\label{tab:oe_table}
\centering
\small
\vspace{1mm}
\setlength{\cmidrulesep}{0pt}
\resizebox{\linewidth}{!}{
\begin{tabular}{@{}l c ll l @{}}
\Xhline{2\arrayrulewidth}
\multirow{2}{*}{\textbf{Model}} & \multirow{2}{*}{\textbf{LLM}} & \multicolumn{2}{c}{\textbf{ActivityNet-QA}} & \multicolumn{1}{c}{\textbf{VCGPT}} \\
\cmidrule(lr){3-4}\cmidrule(lr){5-5}
 &  & \multicolumn{1}{c}{acc} & \multicolumn{1}{c}{score} & \multicolumn{1}{c}{score} \\
\hline
VideoLLaMA3 & \multirow{2}{*}{7B} & \textcolor{gray}{52.4} & \textcolor{gray}{70.2} & \textcolor{gray}{60.3} \\
\hspace{0.5em}+ \textbf{ReFoCUS} &  & 51.8 \gain{-0.6} & 72.2 \gain{2.0} & 62.7 \gain{2.4} \\
\addlinespace[2pt]
LLaVA-OV & \multirow{2}{*}{7B} & \textcolor{gray}{50.9} & \textcolor{gray}{68.8} & \textcolor{gray}{59.9} \\
\hspace{0.5em}+ \textbf{ReFoCUS} &  & 52.3 \gain{1.4} & 69.7 \gain{0.9} & 61.4 \gain{1.5} \\
\addlinespace[2pt]
InternVL3 & \multirow{2}{*}{8B} & \textcolor{gray}{53.7} & \textcolor{gray}{69.4} & \textcolor{gray}{59.4} \\
\hspace{0.5em}+ \textbf{ReFoCUS} &  & 55.4 \gain{1.7} & 70.4 \gain{1.0} & 61.0 \gain{1.6} \\
\addlinespace[2pt]
InternVL3.5 & \multirow{2}{*}{8B} & \textcolor{gray}{50.4} & \textcolor{gray}{67.7} & \textcolor{gray}{59.5} \\
\hspace{0.5em}+ \textbf{ReFoCUS} &  & 52.2 \gain{1.8} & 68.8 \gain{1.1} & 60.8 \gain{1.3} \\
\addlinespace[2pt]
Qwen3-VL & \multirow{2}{*}{8B} & \textcolor{gray}{50.8} & \textcolor{gray}{65.9} & \textcolor{gray}{60.3} \\
\hspace{0.5em}+ \textbf{ReFoCUS} &  & 52.4 \gain{1.6} & 67.2 \gain{1.3} & 62.3 \gain{2.0} \\
\Xhline{2\arrayrulewidth}
\end{tabular}
}
\end{table}
\endgroup

\subsection{Low Entropy Degradation Analysis}
\begin{figure*}[t!]
\centering
\includegraphics[width=0.99\textwidth]{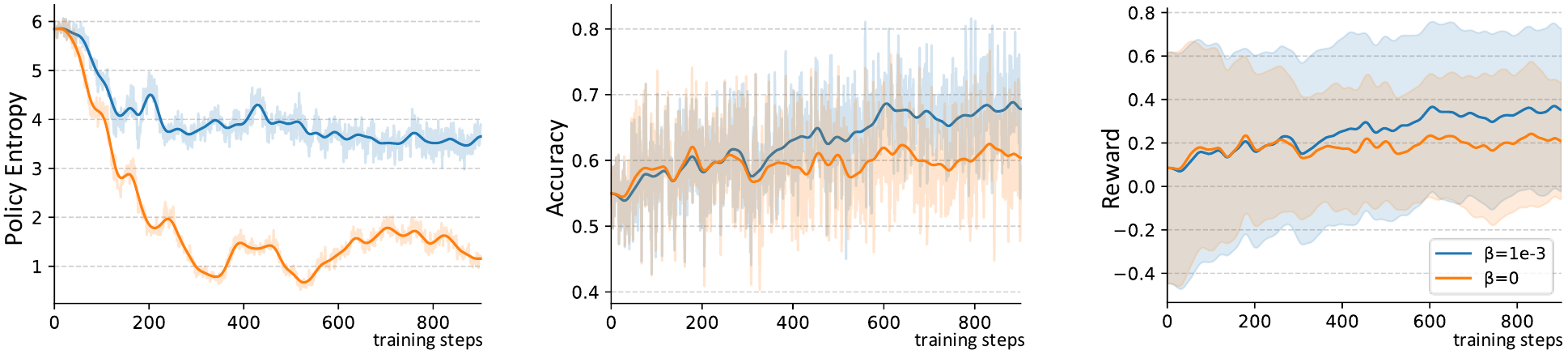}
\vspace{-4mm}
\begin{flushleft}
    \hspace{1.45cm}{(a) Policy Entropy $\mathcal{H}_{\theta}$\hspace{3.2cm}(b) Train Accuracy \hspace{2.8cm}(c) Train reward with $\pm\sigma$}
\end{flushleft}
\vspace{-6mm}
\caption{
Ablation of entropy regularization. Eliminating the entropy coefficient $\beta{=}0$ leads to early entropy collapse and limits long-term performance improvement. Even a small entropy bonus coefficient $\beta$ is sufficient to prevent this degradation.
}
\label{sfig:2}
\end{figure*}

To validate the effect of the entropy bonus $\mathcal{H}(\pi_\theta)$ introduced in \cref{sec:modeling}, we retrained the policy under the Stage\,1 configuration in \cref{tab:train_hparams_stage} with $\beta{=}0$.
Here, $\mathcal{H}(\pi_\theta)$ denotes the expected policy entropy across autoregressive frame-selection steps,
\begin{equation}
\mathcal{H}(\pi_\theta)
= \mathbb{E}_{f_{<i} \sim \pi_\theta}
\!\left[
  \mathcal{H}\bigl(\pi_\theta(\cdot \mid f_{<i}, v, t)\bigr)
\right] ,
\end{equation}
which regularizes the frame-selection distribution to remain sufficiently stochastic and thus preserves the model’s capacity for exploration.  
When the entropy regularization is removed ($\beta{=}0$), the policy entropy $\mathcal{H}(\pi_\theta)$ drops sharply in the early phase, leading to a rapid decline in explorability, the model’s ability to meaningfully search diverse frame combinations, and causing premature convergence (\cref{sfig:2} (a)).  
This degradation manifests as both mean reward and accuracy plateauing at a lower level (\cref{sfig:2} (b)), accompanied by an early collapse of the standard deviation of reward (\cref{sfig:2} (c)).  
Such a reduction in reward variability reflects a narrowing of the performance-improvement horizon, beyond which the policy can no longer discover higher-reward configurations.  
Additionally, the model begins to exhibit false confidence, consistently selecting frames with unwarranted certainty despite limited evidence, reinforcing suboptimal behaviors.  
In contrast, $\beta{=}10^{-3}$ maintains adequate entropy and standard deviation throughout training, preventing false confidence and sustaining a broad performance-improvement horizon.  
This confirms that controlled entropy is essential for maintaining exploration potential while guiding stable convergence toward superior performance.

\subsection{Ablation on Reward Formulation}
\label{sec:reward_ablation}

To validate the necessity of our margin-based reward formulation (\cref{eq:reward}), we compare it against a simpler binary reward variant that assigns a reward of $1$ when the reward model predicts the correct answer and $0$ otherwise. Both variants are trained under identical conditions. As shown in~\cref{tab:reward_ablation}, the binary reward already provides a meaningful improvement over uniform sampling (62.2 $\to$ 64.6), confirming that even coarse feedback is sufficient to guide frame selection toward informative subsets. However, our margin-based design achieves a further gain (64.6 $\to$ 65.4), with the improvement especially pronounced in the medium and long video categories. This indicates that the continuous, magnitude-aware signal captures finer distinctions between frame subsets that the binary indicator cannot express. By encoding how confidently the reward model $r_\varphi$ favors the correct answer over the strongest distractor, the margin reward provides richer gradient information to the policy, enabling it to discriminate between merely adequate and truly optimal frame compositions.

\begin{table}[t]
\caption{
Ablation on reward formulation. We compare our margin-based reward (\cref{eq:reward}) against a binary variant (1 if the reward model predicts the correct answer, 0 otherwise) under the same training setup. We report the mean score of five 7--8B models on Video-MME with 32 frames.
}
\label{tab:reward_ablation}
\centering
\small
\vspace{1mm}
\resizebox{1.0\linewidth}{!}{
\begin{tabular}{@{}lcccc@{}}
\Xhline{2\arrayrulewidth}
\textbf{Reward Design} & \textbf{short} & \textbf{medium} & \textbf{long} & \textbf{overall} \\
\hline
Uniform sampling              & \textcolor{gray}{73.8} & \textcolor{gray}{61.0} & \textcolor{gray}{51.9} & \textcolor{gray}{62.2} \\
\textit{ReFoCUS} (binary)     & \textbf{75.3} & 63.3 & 55.2 & 64.6 \\
\textit{ReFoCUS} (margin, \textbf{ours}) & \textbf{75.3} & \textbf{64.1} & \textbf{56.8} & \textbf{65.4} \\
\Xhline{2\arrayrulewidth}
\end{tabular}
}
\end{table}

\subsection{Ablation on Query Conditioning}
\label{sec:query_ablation}

To quantitatively verify that our policy genuinely conditions its frame selection on the input query rather than learning a fixed, query-agnostic selection pattern, we conduct a \textit{cross-query interference} experiment on Video-MME. Specifically, we replace the correct query with a randomly sampled question from the same video during inference, while keeping all other conditions unchanged. As shown in~\cref{tab:query_ablation}, providing a random question leads to a significant performance drop (65.4 $\to$ 58.1), which falls even below the uniform sampling baseline (62.2). This result demonstrates that the policy actively filters out frames deemed irrelevant to the conditioning query: when guided by an incorrect question, it suppresses the visual evidence required for the actual answer, thereby degrading performance below that of uninformed uniform selection. The finding confirms that query conditioning plays a critical role in guiding effective frame selection and that \textit{ReFoCUS} performs genuine query-conditioned semantic grounding rather than relying on superficial visual saliency or fixed temporal heuristics.

\begin{table}[t]
\caption{
Ablation on query conditioning. We evaluate the effect of providing an incorrect query during frame selection. ``Random Question'' replaces the correct query with a randomly sampled question from the same video. We report the mean score of five 7--8B models on Video-MME with 32 frames.
}
\label{tab:query_ablation}
\centering
\small
\vspace{1mm}
\resizebox{1.0\linewidth}{!}{
\begin{tabular}{@{}lcccc@{}}
\Xhline{2\arrayrulewidth}
\textbf{Query Condition} & \textbf{short} & \textbf{medium} & \textbf{long} & \textbf{overall} \\
\hline
Uniform sampling                     & \textcolor{gray}{73.8} & \textcolor{gray}{61.0} & \textcolor{gray}{51.9} & \textcolor{gray}{62.2} \\
\textit{ReFoCUS} (Random Question)   & 66.0 & 55.1 & 53.1 & 58.1 \\
\textit{ReFoCUS} (Correct Question)  & \textbf{75.3} & \textbf{64.1} & \textbf{56.8} & \textbf{65.4} \\
\Xhline{2\arrayrulewidth}
\end{tabular}
}
\end{table}

\section{Qualitative Case Study}
We present qualitative comparisons across uniform sampling, \textsc{mDP$^3$}~\cite{sun2025mdp3}, and our \textit{ReFoCUS} policy in~\cref{fig:qaexample-assets/039-1,fig:qaexample-assets/206-1,fig:qaexample-assets/345-1,fig:qaexample-assets/419-2,fig:qaexample-assets/479-1,fig:qaexample-assets/618-1}. Each example visualizes the selected frames along the timeline, enabling direct inspection of where each method concentrates its evidence. The uniform sampling selects frames at fixed temporal intervals regardless of the question, and therefore it often fails to include the critical temporal segments required for answering the query. \textsc{mDP$^3$} shows stronger focus than uniform sampling, but its decisions are largely driven by low-level RGB similarity or visually salient fragments rather than query-conditioned semantics, resulting in inconsistent reasoning.

In contrast, \textit{ReFoCUS} consistently selects compact yet semantically aligned frame subsets that directly correspond to the question intent. This query-conditioned behavior is clearly visible in the timestamp visualizations: frames cluster around the temporal regions that contain the decisive visual cues required to answer the question. Notably, for some cases exemplified in~\cref{fig:qaexample-assets/618-1}, \textit{``What do the expanding red lines on the map in the first few minutes of the video stand for?''}, both uniform sampling and \textsc{mDP$^3$} fail to identify or densely sample the early-video interval mentioned in the query. \textit{ReFoCUS}, however, accurately localizes this segment, gathers the relevant evidence concentrated in the first few minutes, and consequently provides the correct answer. These examples demonstrate that the learned policy performs genuine semantic grounding rather than heuristic or appearance-based selection.

\section{Failure Case and Limitation}
Although \textit{ReFoCUS} demonstrates robust query-conditioned frame selection across diverse scenarios, it also has some limitations. As illustrated in~\cref{fig:qaexample-assets/041-2,fig:qaexample-assets/792-1}, since the selection process operates under a fixed frame budget, the policy sometimes uses most of its budget too early in the video. When this happens, it may focus on initial frame segments and miss later events that are still relevant to the question. Such cases mostly appear in long videos that contain multiple separated temporal cues, where early commitment prevents the policy from observing the full temporal context.

While these errors are relatively infrequent, they highlight an inherent trade-off between early commitment and global temporal coverage. Future work may explore more adaptable selection strategies or hierarchical policies that dynamically reallocate the budget based on the observed evidence, potentially mitigating premature focus and improving global temporal awareness.

\subsection{Discussion and Limitation} While \textit{ReFoCUS} introduces a new perspective by shifting policy optimization from output-level textual behavior to input-level visual grounding, several limitations remain. Similar to other RL-based methods, our training requires substantial computational cost due to repeated autoregressive sampling and reward evaluation. Moreover, the learned policy can reflect biases or blind spots of the reward model, since its selection behavior is shaped by the model’s visual scoring patterns. Despite these challenges, our results show that modeling selection behavior directly at the input level can yield semantically coherent and contextually meaningful frame subsets. We believe that this direction opens new opportunities for aligning LMM behavior not only by how they respond, but also by how they perceive and prioritize visual information.

\RenderQAExampleFigure{assets/039-1}{
Qualitative comparison of temporal localization across uniform sampling, \textsc{mDP$^3$}, and \textit{ReFoCUS}.
}
\RenderQAExampleFigure{assets/206-1}{
Qualitative comparison of temporal localization across uniform sampling, \textsc{mDP$^3$}, and \textit{ReFoCUS}.
}
\RenderQAExampleFigure{assets/345-1}{
Qualitative comparison of temporal localization across uniform sampling, \textsc{mDP$^3$}, and \textit{ReFoCUS}.
}
\RenderQAExampleFigure{assets/419-2}{
Qualitative comparison of temporal localization across uniform sampling, \textsc{mDP$^3$}, and \textit{ReFoCUS}.
}
\RenderQAExampleFigure{assets/479-1}{
Qualitative comparison of temporal localization across uniform sampling, \textsc{mDP$^3$}, and \textit{ReFoCUS}.
}
\RenderQAExampleFigure{assets/618-1}{
Qualitative comparison of temporal localization across uniform sampling, \textsc{mDP$^3$}, and \textit{ReFoCUS}.
}


\RenderQAExampleFigure{assets/041-2}{
Comparison of representative failure cases across uniform sampling, \textsc{mDP$^3$}, and \textit{ReFoCUS}.
}
\RenderQAExampleFigure{assets/792-1}{
Comparison of representative failure cases across uniform sampling, \textsc{mDP$^3$}, and \textit{ReFoCUS}.
}


\end{document}